\documentclass[10pt,twocolumn,letterpaper]{article}

\usepackage[pagenumbers]{cvpr}

\definecolor{cvprblue}{rgb}{0.21,0.49,0.74}
\usepackage[pagebackref,breaklinks,colorlinks,allcolors=cvprblue]{hyperref}
\usepackage{algorithm}
\usepackage{algorithmic}
\usepackage{pifont}

\title{EACO: Enhancing Alignment in Multimodal LLMs via Critical Observation}

\author{Yongxin Wang$^{1}$, Meng Cao$^{1}$, Haokun Lin$^{1}$, Mingfei Han$^{1}$, Liang Ma$^{1}$,
\\
Jin Jiang$^{1}$, Yuhao Cheng$^{2}$, Xiaodan Liang$^{1}$\\
$^{1}$ Mohamed bin Zayed University of Artificial Intelligence 
$^{2}$ Lenovo Research 
}

\begin{document}

\twocolumn[{%
\renewcommand\twocolumn[1][]{#1}%
\maketitle

\begin{center}
    \centering
    \captionsetup{type=figure}
      \includegraphics[width=\linewidth]{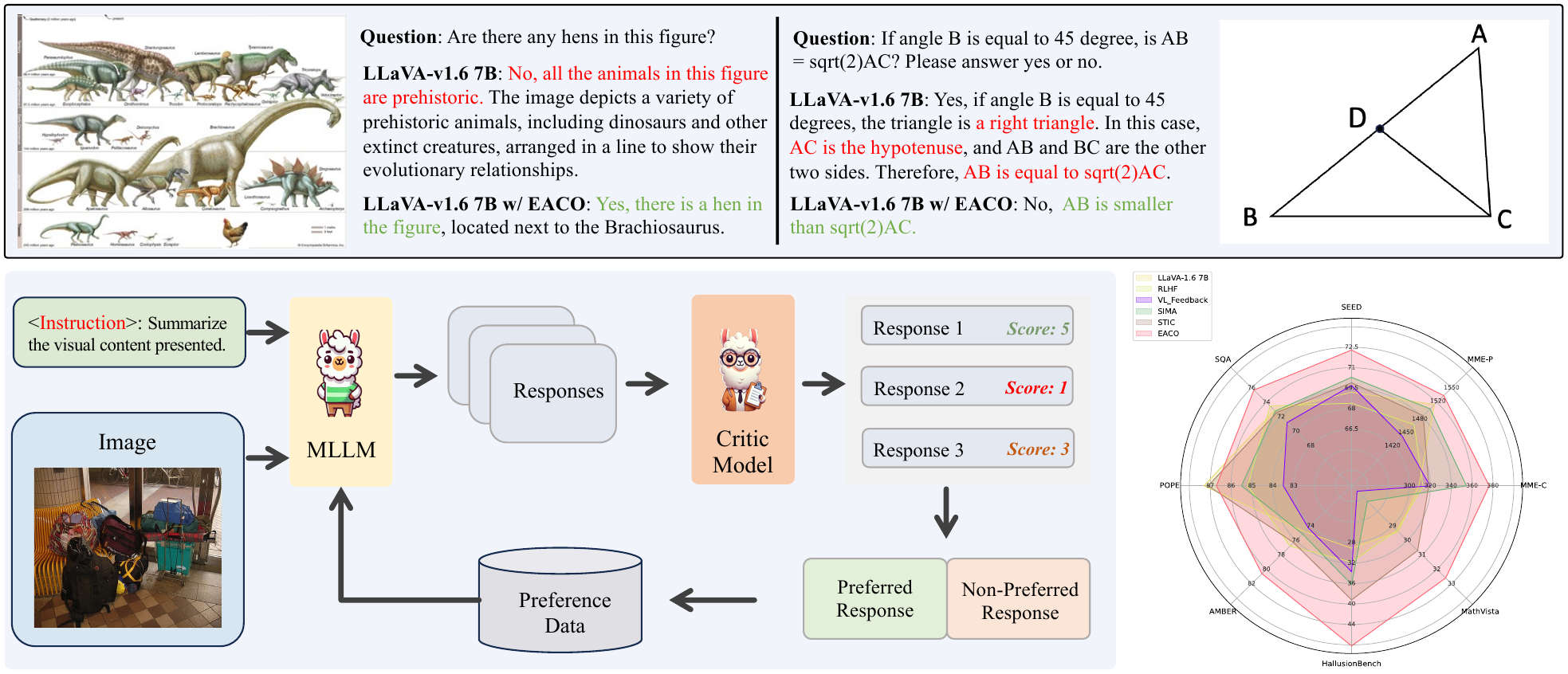}
  \vspace{-2mm}
    \captionof{figure}{\textbf{Upper}: Response examples from original LLaVA-v1.6 7B\cite{liu2023improvedllava} and LLaVA-v1.6 7B w/ EACO, which mitigates hallucination and improve reasonging ability. \textbf{Lower Left}: The framework of EACO. The process begins with an image-question pair, which is fed to the initialized MLLM to genrate multiple responses. And then these responses are evaluated by a Critic Model that provides judgments regarding their quality. Based on the critic's analysis, the responses are categorized into preference and non-preference groups. Finally, preference data is subsequently collected to further improve the MLLMs. \textbf{Lower Right}: Comparison with LLaVA-v1.6 7B\cite{liu2023improvedllava}, LLaVA-RLHF\cite{llava_rlhf}, Silkie\cite{2023vlfeedback}, SIMA\cite{sima}, and STIC\cite{stic}.  Our proposed EACO framework achieves improvements across multiple metrics, demonstrating robust performance gains compared to other methods.}
  \label{fig:pre}
\end{center}%
  \vspace{-1mm}

}]


\begin{abstract}

Multimodal large language models (MLLMs) have achieved remarkable progress on various visual question answering and reasoning tasks  leveraging instruction fine-tuning specific datasets. They can also learn from preference data annotated by human to enhance their reasoning ability and mitigate hallucinations. Most of preference data is generated from the model itself. However, existing methods require high-quality critical labels, which are costly and rely on human or proprietary models like GPT-4V. In this work, we propose \textbf{E}nhancing \textbf{A}lignment in MLLMs via \textbf{C}ritical \textbf{O}bservation (\textbf{EACO}), which aligns MLLMs by self-generated preference data using only 5k images economically. Our approach begins with collecting and refining a Scoring Evaluation Instruction-tuning dataset to train a critical evaluation model, termed the Critic. This Critic observes model responses across multiple dimensions, selecting preferred and non-preferred outputs for refined Direct Preference Optimization (DPO) tuning. To further enhance model performance, we employ an additional supervised fine-tuning stage after preference tuning. EACO reduces the overall hallucinations by 65.6\% on HallusionBench and improves the reasoning ability by 21.8\% on MME-Cognition. EACO achieves an 8.5\% improvement over LLaVA-v1.6-Mistral-7B across multiple benchmarks. Remarkably, EACO also shows the potential critical ability in open-source MLLMs, demonstrating that EACO is a viable path to boost the competence of MLLMs.
\end{abstract}

\section{Introduction}
\label{sec:intro}


    

In recent years, Large Language Models (LLMs) have achieved remarkable success, largely driven by scaling up model size and enhancing data quality. In real-world applications, integrating inputs from other modalities, such as visual and auditory information, has propelled advancements in Multimodal Large Language Models (MLLMs)~\cite{liu2023llava,liu2023improvedllava,blip}. Existing MLLMs have demonstrated notable progress, especially within the research community~\cite{liu2023improvedllava, chen2023internvl, Qwen-VL}, excelling in a range of downstream multimodal tasks such as visual question answering and image captioning~\cite{fu2023mme,lu2024mathvista,cococaption}.

Despite these achievements, MLLMs still face critical challenges in reasoning and hallucination. For example, MLLMs sometimes generate descriptions that include counterfactual visual elements, reflecting inaccuracies in visual reasoning and comprehension~\cite{liu2024survey} in Figure \ref{fig:pre}. To mitigate reasoning errors and hallucinations, ongoing research has focused on several core strategies that enhance model performance across modalities. Creating high-quality annotated datasets for MLLMs~\cite{liu2023llava,li2023llavamed,m3it, lrv} is one of the most popular and effective approaches to improve model accuracy, reduce hallucinations, and enhance reasoning capabilities. However, such method is also highly resource-intensive, often presenting significant costs and logistical challenges. 

Recently, fine-tuning methods like Direct Preference Optimization~\cite{rafailov2024direct} (DPO) refine model responses by aligning them more closely with human preferences. Utilizing preference data to guide the model in selecting preferred responses over less preferred alternatives. This approach discourages outputs that may be factually incorrect or lack logical coherence. While most existing methods~\cite{llava_rlhf,2023vlfeedback} rely on feedback from human annotators or proprietary models to generate preference data, these approaches are often costly and dependent on access to specialized resources~\cite{2023vlfeedback}. 
We seek to address these limitations by developing a scalable, critic-guided preference alignment strategy that uses self-generated preference data, thus reducing reliance on external feedback sources while maintaining high alignment quality. 
One recent pre-print work LLaVA-Critic~\cite{llava_critic} also introduces the similar critic-based learning. The difference lies in that our EACO focuses on the self-enhancement in DPO training while LLaVA-Critic prioritizes building a generalist evaluator. Given that one application of LLaVA-Critic is to provide reward-signals for iterative DPO training, we provide the detailed discussion from the perspectives of data composition, scoring selection, and training objectives in Section \ref{subsec:2_2}.

To enhance reasoning abilities and reduce hallucinations in MLLMs, our work introduces a novel and economical critic-based method. Specifically, we collect and refine a critic dataset comprising 51,000 images and over 137,000 critic instructions. This dataset is designed to fine-tune the model, enabling it to assess and critique responses with greater accuracy. The optimization process starts with a seed model that undergoes training on the critic dataset. This initial model is progressively fine-tuned to develop its critic abilities, allowing it to evaluate responses across multiple dimensions, such as Relevance, Basic Elements, and Clarity. After incorporating critic guidance into the response generation, the Critic then selects preferred and non-preferred outputs, which are used to guide refined Direct Preference Optimization~\cite{rafailov2024direct} (DPO) tuning. The DPO process aligns the model more closely with high-quality responses, reducing tendencies toward hallucination and improving overall reasoning capabilities.

The primary contribution of this work is the development of a novel, critic-based training framework that enhances reasoning abilities and reduces hallucinations in MLLMs in an economical manner. Key contributions include:

\begin{itemize}
    \item We propose \textbf{EACO}, a novel critic-based approach that guides MLLMs toward generating more accurate, contextually relevant, and hallucination-free responses. 
    \item We refine a \textbf{Large-scale Critic Dataset} containing 51,000 images and over 137,000 detailed critique instructions. This dataset is designed to train the model’s “critic” abilities, enabling it to assess responses across multiple dimensions.
    \item EACO achieves an \textbf{8.5\%} average improvement on multiple benchmarks over baseline, which indicates effectiveness of the critic-based framework, especially on reasoning task and hallucination mitigation.
\end{itemize}
\section{Related Work}

\subsection{Multimodal Large Language Models (MLLMs)} MLLMs~\cite{liu2023llava,liu2023improvedllava,chen2023internvl,blip,cao2024continual} have dominated a wide range of multimodal tasks, achieving remarkable progress in vision-language understanding~\cite{liu2023llava,fu2023mme,li2023seed}, reasoning~\cite{lu2024mathvista} and generation~\cite{wang2024emu3}. Most of them rely on pre-trained uni-modal models, where MLLMs use learnable projectors to connect visual encoders with language models. These projectors are typically categorized as either query-based, like Q-Former~\cite{blip} in MiniGPT-4~\cite{zhu2023minigpt} and Instruct-BLIP~\cite{insblip}, which use cross-attention to capture visual signals, or projection layer-based, as seen in LLaVA~\cite{liu2023llava,liu2023improvedllava} and ShareGPT4V~\cite{chen2023sharegpt4v}, where linear projection layers or Multi-Layer Perceptrons (MLPs) map visual features into the input space of language models. In contrast, Fuyu-8B~\cite{fuyu-8b} and Gemini~\cite{team2023gemini} involve end-to-end training without pre-trained components. For example, Fuyu-8B processes raw image patches directly and transforms them into embeddings via linear projection, bypassing the use of pre-trained visual encoders and instead learning the vision-language relationship from scratch.

\begin{figure*}[t]
    \centering
    \begin{minipage}{0.38\textwidth}
        \centering
        \includegraphics[width=\textwidth]{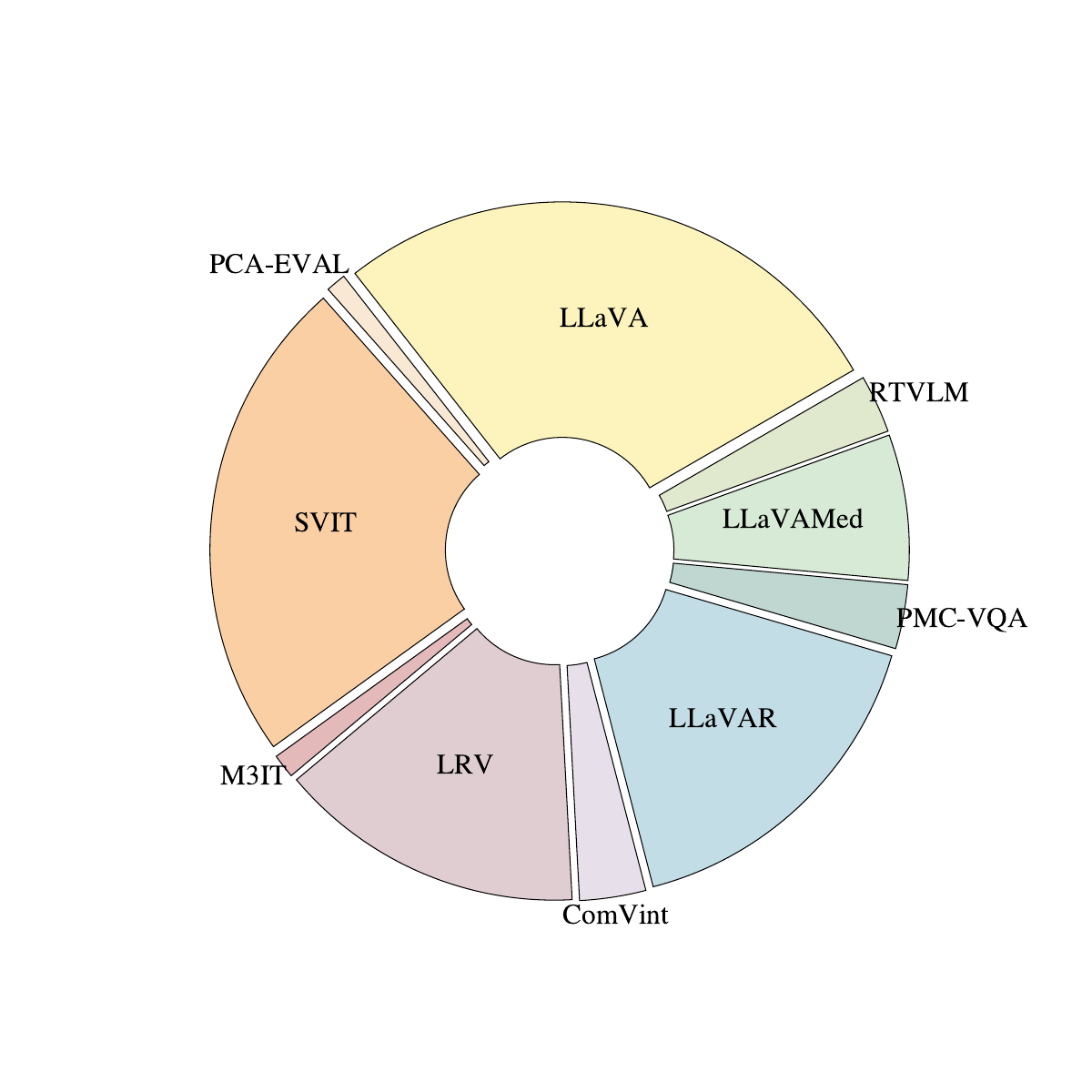}
    \end{minipage}
    \hfill
    \begin{minipage}{0.6\textwidth}
    \small
    \resizebox{\textwidth}{!}{
    \begin{tabular}{l|lr}
    \toprule
    Dataset & Description & \# of Instructions \\
    \midrule
    LLaVA~\cite{liu2023improvedllava}&Visual Instruction Synthesized by GPT-4&14,128\\
    SVIT~\cite{zhao2023svit}& Visual Instruction Synthesized by GPT-4&12,142\\
    LRV~\cite{lrv}&Robust Visual Instruction&7,650\\
    ComVint~\cite{ComVint}&Complex Visual Reasoning Instruction&1,476\\
    LLaVAR~\cite{zhang2023llavar}&Text-rich Image Understanding&8,524\\
    LLaVAMed~\cite{li2023llavamed}&Biomedical Vision-Language Instruction&3,628\\
    PMC-VQA~\cite{zhang2023pmcvqa}&Medical Image Question Answering&1,463\\
    PCA-EVAL~\cite{chen2024pca}&Embodied Decision-making Instruction&246\\
    RTVLM~\cite{li2024red} &Red-Teaming Instructions&1,317\\
    M3IT~\cite{m3it} &Academic Vision-Language Tasks&425\\
    \midrule
    Total &Visual Instructions  & 51,000\\
    Critic Dataset & Scoring Evaluation Instructions & 137,486 \\
    \bottomrule
    \end{tabular}
    }

    \end{minipage}

    \caption{Various datasets used for scoring evaluation. Each dataset contains a specific number of instructions, with a total of 51,000 instructions dedicated to visual tasks and 137,486 for scoring evaluations. Since there are responses with a small score gap in the instructions, we filter out those responses and retain the ones with a larger score gap.}
    \label{tab:critic_data}
\end{figure*}


\subsection{Preference Alignment} \label{subsec:2_2}
Preference alignment~\cite{rafailov2024direct,ppo,instructgpt,meng2024simpo}is a standard technique used in LLMs to strengthen the model's capabilities to improve the instruction-following ability.  The goal of Reinforcement Learning from Human Feedback (RLHF)~\cite{instructgpt} and Proximal Policy Optimization (PPO)~\cite{ppo} is to align language models with user intent across a broad spectrum of tasks by fine-tuning them using human feedback. SimPO~\cite{meng2024simpo} serves as a simpler alternative to the PPO, using a particular choice of reward model parameterization that enables extraction of its optimal policy in closed form, without an RL training loop. LLaVA-RLHF~\cite{llava_rlhf} is the first work in the multi-modal area, which improves visual understanding and instruction-following abilities for MLLMs. Recently, several methods~\cite{2023vlfeedback, csr, yu2024rlaif} have utilized preference data from other AI models to perform preference tuning. While SIMA~\cite{sima} directly uses a self-rewarding mechanism with a fine-grained critic prompt to construct preference data. Both STIC~\cite{stic} and POVID~\cite{povid} introduce the corruption of images to generate non-preference, and then utilize DPO tuning on preference data. 
We acknowledge the recent contributions of LLaVA-Critic \cite{llava_critic} in establishing a generalist reward evaluator and the effective applications in preference learning. We detail the technical difference between our EACO and LLaVA-Critic as follows.

\noindent \textbf{Discussion with LLaVA-Critic:} 
We acknowledge the valuable contributions of LLaVA-Critic~\cite{llava_critic}, including a high-quality critic dataset, multimodal LLMs for evaluation and feedback, and open-source data and codebase. Due to the concern regarding the similarities between LLaVA-Critic and EACO, we outline some key differences.

\noindent \emph{Motivation:} LLaVA-Critic focuses on building a generalist evaluator, while our approach emphasizes self-enhancement of the model. 

\noindent \emph{Data Collection: } We collect data directly from VL-feedback, and after refinement, use distinct templates compared to LLaVA-Critic, including additional information such as total scores.

\noindent \emph{Scoring Selection:} For self-generated responses, LLaVA-Critic adjusts outputs using different hyperparameters, whereas we adopt diverse prompts for generation. Leveraging our critic model's ability to directly output judgments and scores, we adopt a more intuitive selection process, whereas LLaVA-Critic relies on computationally intensive quadratic operations for relative scoring.

\noindent \emph{Training Objective:} For Direct Preference Optimization (DPO), we incorporate two additional regularization terms to ensure the model favors preferred responses and avoids verbose outputs. In contrast, LLaVA-Critic does not elaborate on its DPO training details. By default, we can consider it as traditional DPO~\cite{rafailov2024direct}. Although DPO is an effective tool for preference alignment, it has certain drawbacks, such as generating overly verbose responses. To address this issue, we introduce these modifications to mitigate these problems.

\noindent \emph{Efficiency:} EACO achieves improved performance using less data during the preference alignment stage in Section \ref{exp_setup}. Additionally, to prevent data leakage, we ensured that these data do not overlap with the training data of the critic model.

\noindent \emph{Enhanced SFT:} After preference alignment, EACO introduces an enhanced Supervised Fine-Tuning (SFT) stage to achieve superior performance, different from LLaVA-Critic's pipeline.

Finally, we appreciate the contributions and insights of LLaVA-Critic. Due to the similarity of LLaVA-Critic and our EACO, we provide the above detailed discussions, which is intended to honestly and adequately acknowledge the contributions of existing works. We provide the source code of EACO for reproducibility. We highly value academic integrity and welcome further discussions should there be any additional concerns.



\subsection{Hallucination Mitigation} Hallucination often arises from the model’s attempt to rely on its general knowledge base when it cannot confidently interpret the visual input~\cite{liu2024survey}, leading it to fill gaps with plausible but incorrect information. This can be particularly problematic in applications that require high accuracy, such as medical imaging or autonomous driving. To enhance the perception ability of MLLMs, several works~\cite{liu2023improvedllava, chen2023internvl} scale up the resolution of the vision encoder, while others~\cite{mmvp} have adopted versatile vision encoders, for example,~\cite{mmvp} proposing mixing features from CLIP~\cite{clip} and DINO~\cite{oquab2023dinov2}. There is one more line to mitigate hallucination by introducing the Reinforcement Learning from AI/Human Feedback. LLaVA-RLHF~\cite{llava_rlhf} involves the human feedback to mitigate hallucination by maximizing the human reward. ViGoR~\cite{yan2024vigor} designs a fine-grained reward model,which encompasses both human preferences and automatic metrics, to update the pre-trained model for hallucination mitigation.

\section{Critic Model}

Previous methods often rely on proprietary models, such as GPT-4V, or other large-scale models to build preference datasets, which are crucial for enhancing current Multilingual Language Models (MLLMs)~\cite{2023vlfeedback, yu2024rlaif}. In this section, however, we propose a training pipeline for critic models based on existing MLLMs, using open-source datasets. This approach aims to reduce dependency on proprietary systems while maintaining robust performance.

\subsection{Data Collection}
\label{critic_data_co}

Silkie~\cite{2023vlfeedback} collected over 80,000 multimodal instructions from various datasets~\cite{liu2023improvedllava, zhao2023svit, zhang2023llavar, lrv}, annotated by GPT-4V~\cite{gpt4}. Each instruction includes four responses generated by different models, each evaluated across three dimensions: helpfulness, visual faithfulness, and ethical considerations. For our analysis, we randomly sampled 51,000 of these instructions and refined them for scoring evaluation. To enhance differentiation, we filtered out response pairs with similar scores for the same instructions and images, retaining only those pairs with a larger score gap to improve the robustness of our evaluation.

As shown in Figure \ref{tab:critic_data}, we finally select more than 137k instructions. The refined instruction template is below:
\begin{align*}
    &\texttt{Id, Prompt, Response} \\
    &\texttt{Score:} [ Score_{H} ,Score_{VF}, Score_{EC}] \\
    &\texttt{Total Score}
\end{align*}
where $Score_{H}$, $Score_{VF}$ , and $Score_{EC}$ denote the rating score of response for the image-question pair from perspectives of helpfulness, visual faithfulness, and ethical considerations respectively. And the total score is the sum of these three scores.

\subsection{Critic Model Training}

\begin{figure}
    \centering
    \includegraphics[width=\linewidth]{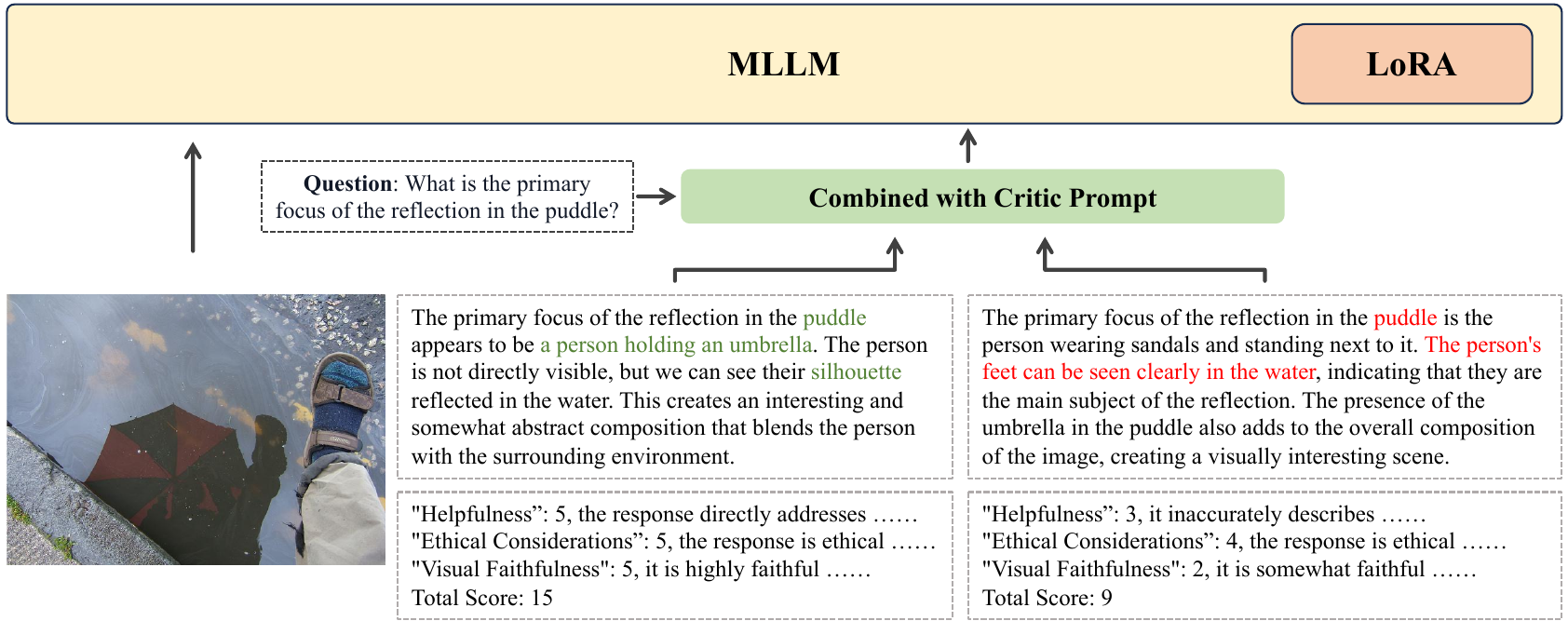}
    \caption{Critic Model training pipeline. After filtering of critic data, we combine the question, responses, and rating score to construct refined critic dataset. Here, we adopt the low-rank adaptatio (LoRA)~\cite{hu2021lora} to fine-tune the Critic Model.}
    \label{fig:training}
\end{figure}

As shown in Figure \ref{fig:training}, after filtering and refining the data, we can proceed to train a specialized critic model for subsequent steps. To achieve this, we fine-tune a pre-trained Multilingual Language Model (MLLM) on the curated scoring evaluation dataset, leveraging its robust capabilities across various downstream tasks. 

Our approach departs from many existing methods that rely on advanced, proprietary models like GPT-4V~\cite{gpt4} or other high-complexity architectures~\cite{yu2024rlaif} for response critique. While these models achieve impressive results, they are often costly, making them challenging to deploy at scale. By fine-tuning a more accessible MLLM, we create a cost-effective critic model that maintains high evaluation accuracy while providing feedback across multiple aspects.

\section{Method: EACO}
\label{methods}

\begin{algorithm}[!ht]
\label{preference_learning}
\caption{EACO: Preference Tuning}
        \begin{algorithmic}
        \STATE \textbf{INPUT:} Unlabeled image dataset: $\left\{\mathbf{v}^{(i)}\right\}_{i \in[N]}$. Image captioning prompt set: $P=\left\{\mathbf{x}^{(i)}\right\}_{i \in\left[M_1\right]}$. MLLM parameterized by $\boldsymbol{\theta}_{0}: p_{\boldsymbol{\theta}_0}$. Critic MLLM parameterized by $\boldsymbol{\theta}^{'}: p_{\boldsymbol{\theta}^{'}}$. Critic Prompt: $\mathbf{x}_{critic}$.

        \FOR{$i=1, \ldots N$}
        \FOR{$j=1, \ldots n$}
        \STATE Randomly sample $\mathbf{x} \sim\left\{\mathbf{x}^{(i)}\right\}_{i \in[M]}$;
        \STATE Generate response $\mathbf{y}_j \sim p_{\boldsymbol{\theta}_t}\left(\cdot \mid \mathbf{v}^{(i)}, \mathbf{x}\right)$;
        \STATE Generate score $S_{i{\mathbf{y}_j}} \sim p_{\boldsymbol{\theta}^{'}}\left(\cdot \mid \mathbf{v}^{(i)}, \mathbf{x}_{critic}\right)$;

        \ENDFOR
        \STATE Select the preferred and non-preferred response $\mathbf{y}_{w} \sim  \arg \max_{j \in [n]} S_{i{\mathbf{y}_j}}$, $\mathbf{y}_{l} \sim  \arg \min_{j \in [n]} S_{i{\mathbf{y}_j}}$;
        \STATE $\operatorname{Add}\left(\mathbf{x}, \mathbf{v}^{(i)}, \mathbf{y}_w, \mathbf{y}_l\right)$ to dataset $D$;
        \ENDFOR
        \STATE Update $\boldsymbol{\theta}_1=\operatorname{argmin}_{\boldsymbol{\theta} \in \boldsymbol{\Theta}} \sum_{\left(\mathbf{x}, \mathbf{v},\mathbf{y}_w, \mathbf{y}_l\right) \in D}
        \newline
        [\log \sigma(\beta \log \frac{p_\theta(y_w \mid x, \mathbf{v})}{p_{\mathrm{ref}}(y_w \mid x, \mathbf{v})} - \beta \log \frac{p_\theta(y_l \mid x, \mathbf{v})}{p_{\mathrm{ref}}(y_l \mid x, \mathbf{v})} 
        \newline
        - \alpha \log p_{\boldsymbol{\theta}}\left(\mathbf{y}_w \mid \mathbf{x}, \mathbf{v}\right)- (\alpha |\mathbf{y}_w| - \alpha |\mathbf{y}_l|))]$;
        \STATE \textbf{OUTPUT:} $\boldsymbol{\theta}_1$.
        \end{algorithmic}
\end{algorithm}

\begin{figure*}
    \centering
    \includegraphics[width=0.96\linewidth]{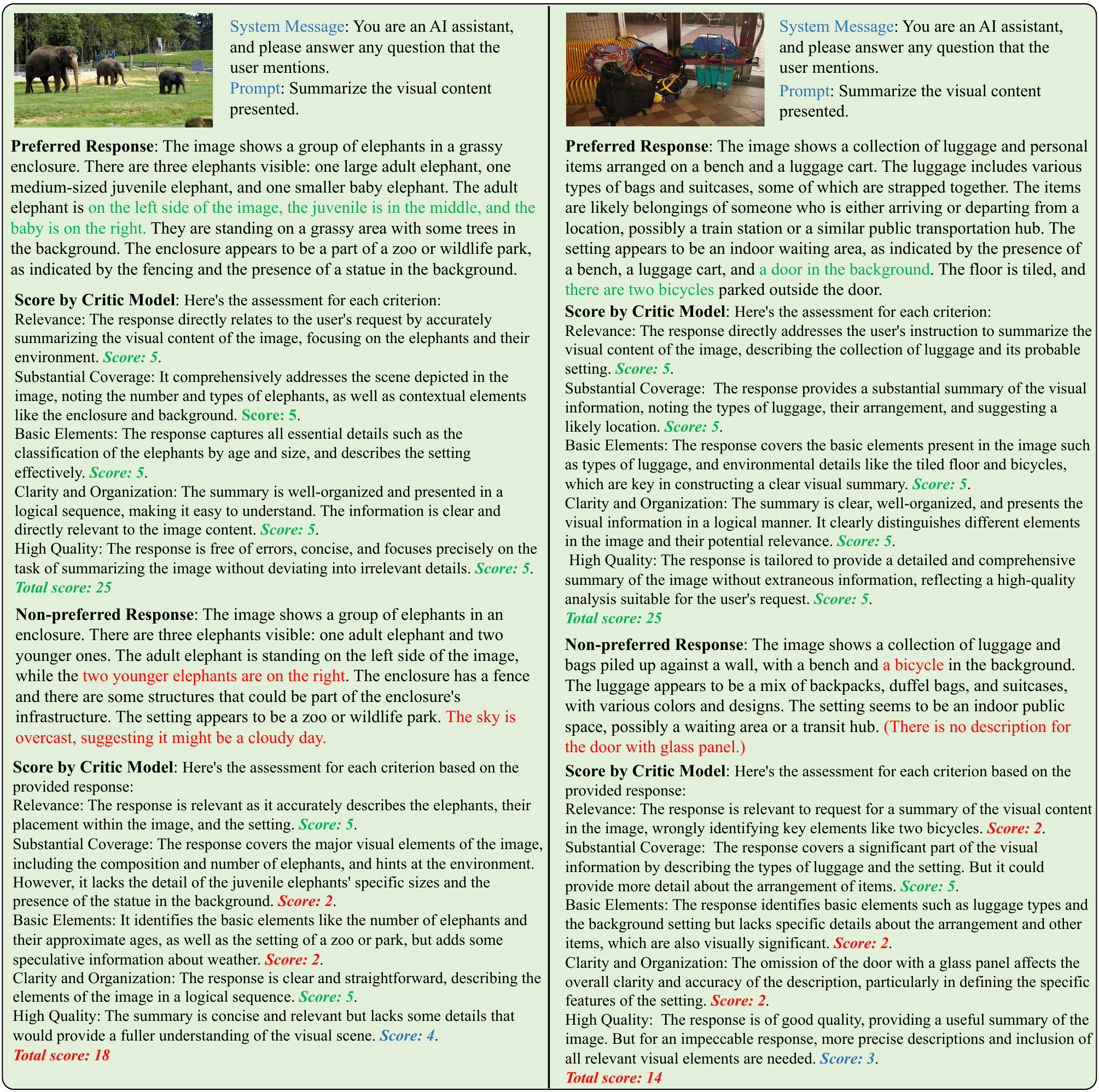}
    \caption{Comparison of preferred and non-preferred responses generated for two visual content summarization examples. The \textbf{left} panel shows a group of elephants in a grassy enclosure, with the preferred response accurately describing the positioning and type of each elephant. The non-preferred response incorrectly describes the arrangement and adds speculative details about the weather. The \textbf{right} panel depicts luggage and personal items in a public waiting area with bicycles in the background. The preferred response correctly identifies key elements and setting details, while the non-preferred response includes inaccurate object descriptions and overlooks the "door".}
    \label{fig:example}
\end{figure*}

In this section, we introduce EACO, a framework composed of three key steps: response generation, critic-guided scoring, and preference learning. In Section \ref{rg}, we first describe the self-generation process, where multiple responses are generated for subsequent selection. Next, Section \ref{cgr} details the critic-guided scoring step, which utilizes a proprietary critic model to evaluate these responses. Finally, in Section \ref{pl}, we explain the model updating process using preference data derived from the scoring step. To further enhance model capabilities, we incorporate an additional supervised fine-tuning (SFT) stage after preference tuning, similar to the approach in~\cite{stic}.

\subsection{Response Generation}
\label{rg}

Most existing methods rely on costly or external models to generate responses for constructing preference datasets, which imposes a substantial computational and financial burden. In contrast, our approach leverages the current Multilingual Language Model (MLLM) to self-generate multiple responses by pairing each image with its corresponding question. As shown in Figure \ref{fig:example}, we obtain Perferred/Non-preferred response pairs after selection.

The generation process begins with a pre-trained MLLM, denoted as $\boldsymbol{\theta}_0$, as the initial checkpoint, along with an unlabeled image dataset ${\mathbf{v}}$ and an image captioning prompt set ${\mathbf{x}}$. For each image-question pair $(\mathbf{v}, \mathbf{x})$, the current MLLM $\boldsymbol{\theta}_0$ generates a set of $n$ responses, $\{y_1, y_2, \dots, y_n\}$.

\subsection{Critic-guided Scoring}
\label{cgr}

After self-generating $n$ responses, we utilize a fine-tuned critic model to evaluate each response using a set of evaluation prompts, producing a score for each response. We then select the responses with the highest and lowest critic scores as the preferred and non-preferred responses, denoted as $\mathbf{y}_w$ and $\mathbf{y}_l$, respectively. These selections are used in the following DPO~\cite{rafailov2024direct} training to further align the current MLLM. The selection process is outlined as follows:

\begin{align}
    \begin{split}
        \mathbf{y}_{w} &=  \arg \max_{j \in [n]} S_{i{\mathbf{y}_{j}}}, \\
        \mathbf{y}_{l} &=  \arg \min_{j \in [n]} S_{i{\mathbf{y}_{j}}}, 
    \end{split}
\end{align}
where $S_{i{\mathbf{y}_{j}}}$ represents the score assigned by the critic model to the $j$-th response in the $i$-th set.

\subsection{Preference Tuning}
\label{pl}

Following the critic-guided scoring step to obtain preference and non-preference pairs, the current MLLM utilizes the preference dataset to conduct preference tuning, utilizing direct preference optimization (DPO)~\cite{rafailov2024direct}. We define the preference dataset as $ D = {\left(\mathbf{x}, \mathbf{v}, \mathbf{y}_w, \mathbf{y}_l\right)} $, where $\mathbf{x}$ is the question, $\mathbf{v}$ is the associated image, $\mathbf{y}_w$ and $\mathbf{y}_l$ denote the preferred and less preferred response respectively. 

To enhance training, we refine the DPO objective function by incorporating two regularization terms: one to reinforce selection of the preferred response and another to discourage verbose responses. The refined DPO objective function is defined as follows:

\begin{align}
    \begin{split}
        &\mathcal{L}\left(p_\theta ; p_{\mathrm{ref}}\right) = -\mathbb{E}_{\left(x, \mathbf{v}, y_{w}, y_{l}\right) \sim \mathcal{D}} \\
        &[\log \sigma(\beta \log \frac{p_\theta(y_{w} \mid x, \mathbf{v})}{p_{\mathrm{ref}}(y_{w} \mid x, \mathbf{v})} - \beta \log \frac{p_\theta(y_{l} \mid x, \mathbf{v})}{p_{\mathrm{ref}}(y_{l} \mid x, \mathbf{v})} \\
        &- \alpha \log p_{\boldsymbol{\theta}}\left(\mathbf{y}_{w} \mid \mathbf{x}, \mathbf{v}\right)- (\alpha |\mathbf{y}_{w}| - \alpha |\mathbf{y}_{l}|))],
    \end{split}
\end{align}

where $p_\theta$ is the current optimized MLLM and $p_{\mathrm{ref}}$ is the reference model, initialized by $p_{\boldsymbol{\theta}_0}$. $\alpha$ is a regularization coefficient, whose value is presented in Suppl. Section\ref{train_details}.

Here, we apply an enhanced Supervised Fine-Tuning (SFT) stage to further optimize model performance, like~\cite{stic} The enhanced approach involves reusing samples from the tuning dataset for additional training. In this setup, the instruction set is enriched by incorporating descriptions generated by the model after Direct Preference Optimization (DPO) tuning. By using these DPO-enhanced descriptions, the SFT phase reinforces alignment with refined model responses, allowing the model to improve consistency in generating high-quality outputs.

\section{Experiments}

\begin{table*}[ht]
    \centering
    \small
    \resizebox{\textwidth}{!}{
    \begin{tabular}{l|cc|cccccccc}
        \toprule
        & \multicolumn{2}{c}{\textbf{Source}} & \multicolumn{3}{c}{\textbf{Comprehensive Benchmark}} & \multicolumn{2}{c}{\textbf{Domain-Specific VQA}} & \multicolumn{3}{c}{\textbf{Hallucination Benchmark}} \\
        \cmidrule(lr){2-3} \cmidrule(lr){4-6} \cmidrule(lr){7-8} \cmidrule(lr){9-11}
        \textbf{Method} &\textbf{Feedback}& \textbf{Response}& \textbf{MME$^P$} & \textbf{MME$^C$} & \textbf{SEED}   & \textbf{SQA$^I$}  & \textbf{MathVista} & \textbf{POPE} & \textbf{AMBER} & \textbf{HallusionBench} \\
        \midrule
        LLaVA-v1.6-7B~\cite{liu2023improvedllava}&\ding{55} &\ding{55}& 1512.3& 308.9& 69.2&72.9&29.4&87.3& 74.8&29.1\\
        + RLHF~\cite{llava_rlhf}& Human&Human&1467.6&312.9 &68.4 &73.5&29.3&85.2& 76.8&31.8\\
        + RLHF-V~\cite{yu2024rlhfv} & Human &Human&1480.1&320.4 & 70.0&72.7&29.7&86.4& 75.7&31.7\\
        + VL-Feedback~\cite{2023vlfeedback} &GPT-4V &GPT-4V&1441.4&319.6 & 69.7&71.2&26.6&83.5&74.3 &33.7\\
        + POVID~\cite{povid}& \ding{55} &Self&1416.7& 349.3&68.9 & 74.3&29.1&87.2&76.2 &37.8\\
        + SIMA~\cite{sima} & Self &Self&1502.8& 354.6&70.3 &72.8&27.3&85.5& 75.1&36.2\\
        + CSR~\cite{csr}& Others &Self& 1520.8&366.8 &70.9 &75.1&28.7&\bf 87.7& 78.4 &35.3\\
        + STIC~\cite{stic}& \ding{55} &Self& 1487.6&325.7 &71.7 &73.1&31.2&83.1&78.9 &48.0\\
        + RLAIF-V~\cite{yu2024rlaif} &Others &Self&1496.4& 318.6& 69.9&72.6&30.7&87.1& 76.3&39.2\\
        + Self-Rewarding~\cite{yuan2024self}&Self&Self&1506.2&319.6&69.9&72.8&29.4&85.4&76.0&30.6 \\
        \textbf{EACO} & Critic &Self&\bf 1532.8 & \bf 376.4 & \bf 72.3& \bf 75.7& \bf 32.6&86.7& \bf 80.6&\bf 48.2\\
        \midrule
        Bunny-8B~\cite{he2024bunny} & \ding{55} &\ding{55}&1636.6&351.1 &73.5 &79.1&34.4&86.9& 77.3&37.7\\
        \textbf{ EACO}  & Critic &Self&\bf 1651.6& \bf 373.9&73.1 &\bf 81.8&\bf 36.8&86.8& \bf 82.9&\bf 49.7\\
        \midrule
        MiniCPM-V-8B~\cite{yao2024minicpm} & \ding{55} &\ding{55}&1663.4&565.7 &74.2 &96.7&61.3&83.2& 80.4&58.7\\

         \textbf{EACO} & Critic &Self&1658.8&\bf 572.1 &\bf 75.1 &96.3&\bf 61.8& \bf 85.7& \bf 83.1 &\bf 60.3\\
        \bottomrule
    \end{tabular}
    }
    \caption{Comparison of performance metrics across various benchmarks for different methods. All the methods are reproduced based on LLaVA-v1.6-7B and evaluated by~\cite{duan2024vlmevalkit} for fairness. We emphasize the source of feedback and response, "Ohters" denotes the feedback from other stronger models, "Self" means self-generated response. MME$^P$ and MME$^C$ denote perception and cognition task of MME benchmark, respectively. SQA$^I$ denote Image part of ScienceQA.}
    \label{tab:performance_comparison}
\end{table*}


Building on our previous work, this section explores preference alignment tuning using Direct Preference Optimization (DPO) with the assistance of a fine-tuned critic model. We begin by outlining the experimental setups in Section \ref{exp_setup}, including baseline methods and training details. Next, in Section \ref{exp_eval}, we describe the benchmarks used and present the main results along with a detailed analysis. Finally, Section \ref{ablation} provides an ablation study to assess the impact of different components in our approach.

\subsection{Experimental Setup}
\label{exp_setup}


\textbf{Implementation Details.} In our experiments, we use LLaVA-v1.6-Mistral-7B~\cite{liu2023improvedllava}, Bunny-8B~\cite{he2024bunny}, and MiniCPM-V2.6 8B~\cite{yao2024minicpm} as backbone models. For critic model training, we employ low-rank adaptation (LoRA) fine-tuning~\cite{hu2021lora} on the \textbf{137k} refined dataset described in Section \ref{critic_data_co}. Detailed training hyperparameters are provided in Suppl. Section \ref{train_details}. To create the self-generated preference dataset, we randomly select \textbf{5k} unlabeled images from the MSCOCO dataset~\cite{cococaption}, following the optimization process outlined in Section \ref{methods} for preference tuning. After preference learning, we apply an enhanced supervised fine-tuning (SFT) stage, inspired by~\cite{stic}. For this SFT stage, we randomly subsample \textbf{5k} instruction-tuning data from the models' original SFT datasets, constructing fine-tuning data with model-generated image captions. Here, we also adopt LoRA fine-tuning for efficient training. Importantly, we ensure there is no overlap between the images in the preference data, enhanced SFT data, and the critic data by checking and filtering out any duplicates. The total tuning cost of EACO is about 2.6 GPU-hours on A100 40G machine when using LLaVA-v1.6-Mistral-7B.

\textbf{Baseline.} Apart from three baseline models mentioned above, we will compare EACO with several preference data-driven methods, including Silkie (VL-Feedback)~\cite{2023vlfeedback}, LLaVA-RLHF~\cite{llava_rlhf},RLHF-V~\cite{yu2024rlhfv}, POVID~\cite{povid}, SIMA~\cite{sima}, CSR~\cite{csr}, STIC~\cite{stic}, and RLAIF-V~\cite{yu2024rlaif}. These comparisons allow us to evaluate the performance of EACO against a range of state-of-the-art preference alignment methods, highlighting its effectiveness in enhancing reasoning ability and reducing hallucinations.

\textbf{Benchmark Evaluation.} We consider the evaluation of models from three aspects, including comprehensive benchmark, domain-specific VQA, and hallucination benchmark. To access the overall capability of MLLMs, we adopt two comprehensive benchmarks, MME~\cite{fu2023mme} and SEEDBench~\cite{li2023seed}. Moreover, we perform evaluation on two domain-specific VQA datasets, ScienceQA~\cite{sciqa} and MathVista~\cite{lu2024mathvista},for domain-specific capability assessment. We additionally evaluated the extent of hallucinations exhibited by models on POPE~\cite{pope}, AMBER~\cite{amber}, and HallusionBench~\cite{guan2023hallusionbench}.

\subsection{Quantitative Results and Analysis}
\label{exp_eval}

The quantitative results presented in Table \ref{tab:performance_comparison} reveals significant insights into the comparative performance of various visual language models across multiple benchmarks. Our proposed EACO consistently demonstrates an improvement in performance compared to other methods, suggesting that incorporating a Critic mechanism along with self-generated responses yields robust advancements across different evaluation metrics.


For the LLaVA-v1.6-Mistral-7B model, incorporating EACO leads to substantial gains across multiple benchmarks. In particular, EACO achieves remarkable results on the comprehensive benchmarks MME$^P$ and MME$^C$, scoring 1532.8 and 376.4, respectively. These scores surpass all other methods, showcasing enhanced perception and cognitive capabilities within the model. Notably, our method also improves performance on both SEED and MathVista, with scores of 72.3 and 32.6, respectively. This improvement reflects an enhanced ability to understand and reason across both general and domain-specific visual question answering (VQA) tasks. In the hallucination benchmarks, our approach attains state-of-the-art scores on AMBER and HallusionBench, with scores of 80.6 and 48.2, respectively. These results indicate a significant reduction in model hallucinations, reinforcing EACO’s effectiveness in controlling false or unsupported content generation.

For the Bunny-8B model, incorporating our approach yields improvements across nearly all metrics. Notably, scores on the AMBER and HallusionBench benchmarks reach 82.9 and 49.7, respectively, indicating a significant reduction in model hallucinations. Additionally, the MME$^P$ score increases from 1636.6 to 1651.6, highlighting the effectiveness of EACO in enhancing the model's perceptual understanding. The improved SQA$^I$ score of 81.8 further demonstrates enhanced performance in domain-specific VQA, particularly in the science domain, showcasing the model’s refined reasoning and domain-specific capabilities.

For the MiniCPM-V-8B model, our approach yields substantial improvements across multiple metrics. Notably, the MME$^C$ score rises from 565.7 to 572.1, underscoring enhanced cognitive abilities. The model also exhibits notable gains on the MathVista, AMBER, and HallusionBench benchmarks, with scores of 61.8, 83.1, and 60.3, respectively. These results highlight the scalability of EACO across different models, demonstrating its effectiveness in reducing hallucinations and improving performance on domain-specific tasks.

\textbf{EACO is a scalable method to enhance MLLMs.} EACO consistently boosts performance across a range of benchmarks and model architectures, proving its scalability and adaptability for various multimodal tasks. This versatility makes it a valuable framework for optimizing diverse MLLMs in both general and specialized domains.

\textbf{EACO mitigates hallucination in MLLMs.} By leveraging a critic mechanism, EACO effectively reduces hallucinations across multiple models, achieving state-of-the-art scores on hallucination benchmarks such as AMBER and HallusionBench. This highlights its ability to improve response reliability by minimizing unsupported or fabricated content.

\textbf{EACO help MLLMs improve reasoning ability.} EACO strengthens both general and domain-specific reasoning skills, particularly in visual question answering (VQA) tasks. This is evidenced by performance gains on benchmarks like MME, SEED, and MathVista, indicating enhanced understanding and contextual accuracy.

Overall, our proposed EACO framework consistently achieves improvements across multiple metrics, demonstrating robust performance gains compared to other methods, regardless of the underlying model architecture. These results validate EACO’s effectiveness in enhancing reasoning abilities, reducing hallucinations, and improving domain-specific task performance across various MLLMs. The consistent gains highlight the versatility and generalizability of our approach, making it a valuable tool for optimizing MLLMs across diverse tasks and settings.


\subsection{Ablation Study}
\label{ablation}

To further analyze the impact of our approach, we conducted three ablation studies based on the LLaVA-v1.6-Mistral-7B model: \textbf{(1)}. Scaling Up Data Quantity in preference tuning,
\textbf{(2)}. Iterative Alignment Performance,
\textbf{(3)}. Impact of Different Critic Prompts, and
\textbf{(4)}. Imapct of Critic Model.

To simplify the performance comparison process, we set the maximum score for each benchmark to 100 and take the average score of 7 benchmarks as the model's overall average performance in this section.

\begin{figure}[h]

    \centering
    \begin{minipage}{0.225\textwidth}
        \centering
        \includegraphics[width=\linewidth]{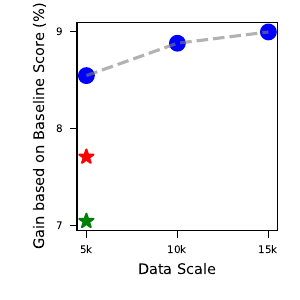}
    \end{minipage}
    \hfill
    \begin{minipage}{0.225\textwidth}
        \centering
        \includegraphics[width=\linewidth]{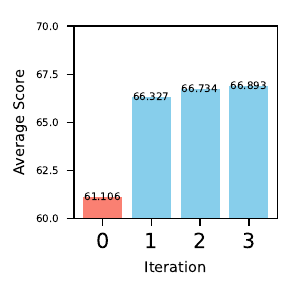}
    \end{minipage}
    \caption{Ablation Studies on Preference Dataset Scaling, Critic Prompt Design, and Iterative Alignment Tuning. \textbf{Left}: The impact of scaling up the preference dataset is shown, where expanding from 5k to 15k samples improves model performance. The figure also compares three different critic prompts: Rating Prompt (the first blue point), Additive Prompt (red star), and Subtractive Prompt (green star). \textbf{Right}: The impact of iterative alignment on model performance, with average scores.}
    \label{fig:ablation}

\end{figure}

\textbf{Scaling up dateset.} We explore the performance when scaling up the preference dataset. The results in Figure \ref{fig:ablation} indicate that scaling up the preference dataset leads to performance improvements, but these improvements exhibit diminishing returns as the dataset grows. Initially, the model obtain a notable gain of about 8.5\% to 66.327 after performing preference tuning on 5k samples. Further increases to 10k and 15k samples result in smaller incremental gains of 5.425 and 5.496 respectively, with improvements becoming marginal—only 0.204 units from 5k to 10k and 0.071 units from 10k to 15k. This trend suggests that while expanding the dataset is beneficial up to a certain point, the value of additional data diminishes as the dataset size increases, potentially indicating an optimal dataset size for balancing model performance and computational efficiency.

\textbf{Iterative Alignment.} The results presented in Figure \ref{fig:ablation} demonstrate the impact of iterative refinement on model performance, as measured by the average score. Starting with a baseline score of 61.106 (Iteration 0), a substantial improvement is observed in Iteration 1, which yields an average score of 66.327. This significant increase indicates the effectiveness of incorporating initial preference tuning. In subsequent iterations, the average score shows marginal improvements, with Iteration 2 reaching 66.734 and Iteration 3 further advancing to 66.893. These results suggest that while the first iteration led to major performance gains, the subsequent iterations achieved more incremental improvements. The diminishing returns after Iteration 1 imply that the model may have reached a plateau in performance, with each additional iteration contributing smaller enhancements. Overall, the results highlight that iterative fine-tuning can effectively boost model performance, particularly in the initial stages, though gains may diminish as more iterations are performed.


\textbf{Critic Prompt.} We also explore the impact of different prompts in the Critic-guided Scoring stage, applying three different critic prompts shown in Suppl. Section \ref{cri_prompt}. The results are shown in Figure \ref{fig:ablation}  marked with the first blue point (Rating Prompt), red star (Additive Prompt), and green star (Subtractive Prompt). This Rating Prompt focuses on multiple dimensions, such as relevance, coverage, clarity, and quality, allowing for a balanced assessment of the responses. And the best result suggests that it can score the model's responses comprehensively and effectively. The Additive Prompt rewards model incrementally for meeting various criteria. While it is less effective in recognizing nuanced response quality. The Subtractive Prompt results in the lowest average score. Compared to the Rating Prompt, it deducts points for any shortcomings, thus making it more challenging for responses to achieve a high score, which affects preference tuning.

\textbf{Self-Rewarding Model.} We conduct some experiments without the assistance of fine-tuned critic models, relying solely on the model itself. The model itself directly critiques self-generated responses with the critic prompt, and then we select and construct the preference dataset. As presented in Table \ref{tab:performance_comparison}, compared to the baseline model, the self-rewarding method shows some minor improvements in certain metrics, such as MME$^P$ improving from 1512.3 to 1506.2 and SEED improving from 69.2 to 69.9.  However, these gains are relatively limited. In contrast, EACO comprehensively outperforms the self-rewarding approach across all evaluation benchmarks, demonstrating the significant importance of critic models. That indicates that the introduction of the critic model

\section{Conclusion}


Our proposed EACO framework offers a scalable and efficient method for guiding MLLMs towards responses that are more accurate, contextually relevant, and largely free of hallucinations. EACO demonstrates generalizability across different model architectures and benchmarks, effectively enhancing both reasoning abilities and reliability in various multimodal applications. This work underscores the potential of critic-based preference alignment as a pathway for optimizing multimodal models to better meet real-world, human-centric needs.


Despite its strengths, EACO’s critic capabilities are largely confined to straightforward tasks such as captioning and basic visual question answering (VQA) tasks. For more complex tasks that require chain-of-thought (CoT) reasoning~\cite{wei2022chain}, the critic model does not yet achieve performance on par with state-of-the-art language models like those in the GPT series~\cite{gpt4}. Nevertheless, we believe that EACO has substantial potential for advancing multimodal understanding and critique capabilities.



{
    \small
    \bibliographystyle{ieeenat_fullname}
    \bibliography{main}

\begin{thebibliography}{54}
\providecommand{\natexlab}[1]{#1}
\providecommand{\url}[1]{\texttt{#1}}
\expandafter\ifx\csname urlstyle\endcsname\relax
  \providecommand{\doi}[1]{doi: #1}\else
  \providecommand{\doi}{doi: \begingroup \urlstyle{rm}\Url}\fi

\bibitem[Achiam et~al.(2023)Achiam, Adler, Agarwal, Ahmad, Akkaya, Aleman, Almeida, Altenschmidt, Altman, Anadkat, et~al.]{gpt4}
Josh Achiam, Steven Adler, Sandhini Agarwal, Lama Ahmad, Ilge Akkaya, Florencia~Leoni Aleman, Diogo Almeida, Janko Altenschmidt, Sam Altman, Shyamal Anadkat, et~al.
\newblock Gpt-4 technical report.
\newblock \emph{arXiv preprint arXiv:2303.08774}, 2023.

\bibitem[Bai et~al.(2023)Bai, Bai, Yang, Wang, Tan, Wang, Lin, Zhou, and Zhou]{Qwen-VL}
Jinze Bai, Shuai Bai, Shusheng Yang, Shijie Wang, Sinan Tan, Peng Wang, Junyang Lin, Chang Zhou, and Jingren Zhou.
\newblock Qwen-vl: A versatile vision-language model for understanding, localization, text reading, and beyond.
\newblock \emph{arXiv preprint arXiv:2308.12966}, 2023.

\bibitem[Bavishi et~al.(2023)Bavishi, Elsen, Hawthorne, Nye, Odena, Somani, and Ta\c{s}\i{}rlar]{fuyu-8b}
Rohan Bavishi, Erich Elsen, Curtis Hawthorne, Maxwell Nye, Augustus Odena, Arushi Somani, and Sa\u{g}nak Ta\c{s}\i{}rlar.
\newblock Introducing our multimodal models, 2023.

\bibitem[Cao et~al.(2024)Cao, Liu, Liu, Wang, Dong, Ding, Zhang, Reid, and Liang]{cao2024continual}
Meng Cao, Yuyang Liu, Yingfei Liu, Tiancai Wang, Jiahua Dong, Henghui Ding, Xiangyu Zhang, Ian Reid, and Xiaodan Liang.
\newblock Continual llava: Continual instruction tuning in large vision-language models.
\newblock \emph{arXiv preprint arXiv:2411.02564}, 2024.

\bibitem[Chen et~al.(2023{\natexlab{a}})Chen, Li, Dong, Zhang, He, Wang, Zhao, and Lin]{chen2023sharegpt4v}
Lin Chen, Jinsong Li, Xiaoyi Dong, Pan Zhang, Conghui He, Jiaqi Wang, Feng Zhao, and Dahua Lin.
\newblock Sharegpt4v: Improving large multi-modal models with better captions.
\newblock \emph{arXiv preprint arXiv:2311.12793}, 2023{\natexlab{a}}.

\bibitem[Chen et~al.(2024)Chen, Zhang, Ren, Zhao, Cai, Wang, Wang, Meng, Liu, and Chang]{chen2024pca}
Liang Chen, Yichi Zhang, Shuhuai Ren, Haozhe Zhao, Zefan Cai, Yuchi Wang, Peiyi Wang, Xiangdi Meng, Tianyu Liu, and Baobao Chang.
\newblock Pca-bench: Evaluating multimodal large language models in perception-cognition-action chain.
\newblock \emph{arXiv preprint arXiv:2402.15527}, 2024.

\bibitem[Chen et~al.(2015)Chen, Fang, Lin, Vedantam, Gupta, Dollar, and Zitnick]{cococaption}
Xinlei Chen, Hao Fang, Tsung-Yi Lin, Ramakrishna Vedantam, Saurabh Gupta, Piotr Dollar, and C.~Lawrence Zitnick.
\newblock Microsoft coco captions: Data collection and evaluation server.
\newblock \emph{arXiv preprint arXiv:1504.00325}, 2015.

\bibitem[Chen et~al.(2023{\natexlab{b}})Chen, Wu, Wang, Su, Chen, Xing, Zhong, Zhang, Zhu, Lu, Li, Luo, Lu, Qiao, and Dai]{chen2023internvl}
Zhe Chen, Jiannan Wu, Wenhai Wang, Weijie Su, Guo Chen, Sen Xing, Muyan Zhong, Qinglong Zhang, Xizhou Zhu, Lewei Lu, Bin Li, Ping Luo, Tong Lu, Yu Qiao, and Jifeng Dai.
\newblock Internvl: Scaling up vision foundation models and aligning for generic visual-linguistic tasks.
\newblock \emph{arXiv preprint arXiv:2312.14238}, 2023{\natexlab{b}}.

\bibitem[Dai et~al.(2023)Dai, Li, Li, Tiong, Zhao, Wang, Li, Fung, and Hoi]{insblip}
Wenliang Dai, Junnan Li, Dongxu Li, Anthony Meng~Huat Tiong, Junqi Zhao, Weisheng Wang, Boyang Li, Pascale Fung, and Steven Hoi.
\newblock Instructblip: Towards general-purpose vision-language models with instruction tuning.
\newblock \emph{arXiv preprint arXiv:2305.06500}, 2023.

\bibitem[Deng et~al.(2024)Deng, Lu, Yin, Hu, Shen, Zou, Chang, and Wang]{stic}
Yihe Deng, Pan Lu, Fan Yin, Ziniu Hu, Sheng Shen, James Zou, Kai-Wei Chang, and Wei Wang.
\newblock Enhancing large vision language models with self-training on image comprehension.
\newblock \emph{arXiv preprint arXiv:2405.19716}, 2024.

\bibitem[Du et~al.(2023)Du, Guo, Zhou, Zhao, Wang, Wang, Cai, Song, and Wen]{ComVint}
Yifan Du, Hangyu Guo, Kun Zhou, Wayne~Xin Zhao, Jinpeng Wang, Chuyuan Wang, Mingchen Cai, Ruihua Song, and Ji-Rong Wen.
\newblock What makes for good visual instructions? synthesizing complex visual reasoning instructions for visual instruction tuning.
\newblock \emph{arXiv preprint arXiv:2311.01487}, 2023.

\bibitem[Duan et~al.(2024)Duan, Yang, Qiao, Fang, Chen, Liu, Dong, Zang, Zhang, Wang, et~al.]{duan2024vlmevalkit}
Haodong Duan, Junming Yang, Yuxuan Qiao, Xinyu Fang, Lin Chen, Yuan Liu, Xiaoyi Dong, Yuhang Zang, Pan Zhang, Jiaqi Wang, et~al.
\newblock Vlmevalkit: An open-source toolkit for evaluating large multi-modality models.
\newblock In \emph{Proceedings of the 32nd ACM International Conference on Multimedia}, pages 11198--11201, 2024.

\bibitem[Fu et~al.(2023)Fu, Chen, Shen, Qin, Zhang, Lin, Yang, Zheng, Li, Sun, et~al.]{fu2023mme}
Chaoyou Fu, Peixian Chen, Yunhang Shen, Yulei Qin, Mengdan Zhang, Xu Lin, Jinrui Yang, Xiawu Zheng, Ke Li, Xing Sun, et~al.
\newblock Mme: A comprehensive evaluation benchmark for multimodal large language models.
\newblock \emph{arXiv preprint arXiv:2306.13394}, 2023.

\bibitem[Guan et~al.(2023)Guan, Liu, Wu, Xian, Li, Liu, Wang, Chen, Huang, Yacoob, et~al.]{guan2023hallusionbench}
Tianrui Guan, Fuxiao Liu, Xiyang Wu, Ruiqi Xian, Zongxia Li, Xiaoyu Liu, Xijun Wang, Lichang Chen, Furong Huang, Yaser Yacoob, et~al.
\newblock Hallusionbench: An advanced diagnostic suite for entangled language hallucination and visual illusion in large vision-language models.
\newblock \emph{arXiv preprint arXiv:2310.14566}, 2023.

\bibitem[He et~al.(2024)He, Liu, Wu, Yuan, Wang, Huang, and Zhao]{he2024bunny}
Muyang He, Yexin Liu, Boya Wu, Jianhao Yuan, Yueze Wang, Tiejun Huang, and Bo Zhao.
\newblock Efficient multimodal learning from data-centric perspective.
\newblock \emph{arXiv preprint arXiv:2402.11530}, 2024.

\bibitem[Hu et~al.(2021)Hu, Shen, Wallis, Allen-Zhu, Li, Wang, Wang, and Chen]{hu2021lora}
Edward~J Hu, Yelong Shen, Phillip Wallis, Zeyuan Allen-Zhu, Yuanzhi Li, Shean Wang, Lu Wang, and Weizhu Chen.
\newblock Lora: Low-rank adaptation of large language models.
\newblock \emph{arXiv preprint arXiv:2106.09685}, 2021.

\bibitem[Li et~al.(2023{\natexlab{a}})Li, Wang, Wang, Ge, Ge, and Shan]{li2023seed}
Bohao Li, Rui Wang, Guangzhi Wang, Yuying Ge, Yixiao Ge, and Ying Shan.
\newblock Seed-bench: Benchmarking multimodal llms with generative comprehension.
\newblock \emph{arXiv preprint arXiv:2307.16125}, 2023{\natexlab{a}}.

\bibitem[Li et~al.(2023{\natexlab{b}})Li, Wong, Zhang, Usuyama, Liu, Yang, Naumann, Poon, and Gao]{li2023llavamed}
Chunyuan Li, Cliff Wong, Sheng Zhang, Naoto Usuyama, Haotian Liu, Jianwei Yang, Tristan Naumann, Hoifung Poon, and Jianfeng Gao.
\newblock Llava-med: Training a large language-and-vision assistant for biomedicine in one day.
\newblock \emph{arXiv preprint arXiv:2306.00890}, 2023{\natexlab{b}}.

\bibitem[Li et~al.(2023{\natexlab{c}})Li, Li, Savarese, and Hoi]{blip}
Junnan Li, Dongxu Li, Silvio Savarese, and Steven Hoi.
\newblock Blip-2: Bootstrapping language-image pre-training with frozen image encoders and large language models.
\newblock \emph{arXiv preprint arXiv:2301.12597}, 2023{\natexlab{c}}.

\bibitem[Li et~al.(2023{\natexlab{d}})Li, Xie, Li, Chen, Wang, Chen, Yang, Wang, and Kong]{2023vlfeedback}
Lei Li, Zhihui Xie, Mukai Li, Shunian Chen, Peiyi Wang, Liang Chen, Yazheng Yang, Benyou Wang, and Lingpeng Kong.
\newblock Silkie: Preference distillation for large visual language models.
\newblock \emph{arXiv preprint arXiv:2312.10665}, 2023{\natexlab{d}}.

\bibitem[Li et~al.(2023{\natexlab{e}})Li, Yin, Li, Chen, Wang, Ren, Li, Yang, Xu, Sun, et~al.]{m3it}
Lei Li, Yuwei Yin, Shicheng Li, Liang Chen, Peiyi Wang, Shuhuai Ren, Mukai Li, Yazheng Yang, Jingjing Xu, Xu Sun, et~al.
\newblock M3it: A large-scale dataset towards multi-modal multilingual instruction tuning.
\newblock \emph{arXiv preprint arXiv:2306.04387}, 2023{\natexlab{e}}.

\bibitem[Li et~al.(2024)Li, Li, Yin, Ahmed, Liu, and Liu]{li2024red}
Mukai Li, Lei Li, Yuwei Yin, Masood Ahmed, Zhenguang Liu, and Qi Liu.
\newblock Red teaming visual language models.
\newblock \emph{arXiv preprint arXiv:2401.12915}, 2024.

\bibitem[Li et~al.(2023{\natexlab{f}})Li, Du, Zhou, Wang, Zhao, and Wen]{pope}
Yifan Li, Yifan Du, Kun Zhou, Jinpeng Wang, Wayne~Xin Zhao, and Ji-Rong Wen.
\newblock Evaluating object hallucination in large vision-language models.
\newblock \emph{arXiv preprint arXiv:2305.10355}, 2023{\natexlab{f}}.

\bibitem[Liu et~al.(2023{\natexlab{a}})Liu, Lin, Li, Wang, Yacoob, and Wang]{lrv}
Fuxiao Liu, Kevin Lin, Linjie Li, Jianfeng Wang, Yaser Yacoob, and Lijuan Wang.
\newblock Aligning large multi-modal model with robust instruction tuning.
\newblock \emph{arXiv preprint arXiv:2306.14565}, 2023{\natexlab{a}}.

\bibitem[Liu et~al.(2023{\natexlab{b}})Liu, Li, Li, and Lee]{liu2023improvedllava}
Haotian Liu, Chunyuan Li, Yuheng Li, and Yong~Jae Lee.
\newblock Improved baselines with visual instruction tuning.
\newblock \emph{arXiv preprint arXiv:2310.03744}, 2023{\natexlab{b}}.

\bibitem[Liu et~al.(2023{\natexlab{c}})Liu, Li, Wu, and Lee]{liu2023llava}
Haotian Liu, Chunyuan Li, Qingyang Wu, and Yong~Jae Lee.
\newblock Visual instruction tuning.
\newblock \emph{NeurIPS}, 2023{\natexlab{c}}.

\bibitem[Liu et~al.(2024)Liu, Xue, Chen, Chen, Zhao, Wang, Hou, Li, and Peng]{liu2024survey}
Hanchao Liu, Wenyuan Xue, Yifei Chen, Dapeng Chen, Xiutian Zhao, Ke Wang, Liping Hou, Rongjun Li, and Wei Peng.
\newblock A survey on hallucination in large vision-language models.
\newblock \emph{arXiv preprint arXiv:2402.00253}, 2024.

\bibitem[Lu et~al.(2022)Lu, Mishra, Xia, Qiu, Chang, Zhu, Tafjord, Clark, and Kalyan]{sciqa}
Pan Lu, Swaroop Mishra, Tony Xia, Liang Qiu, Kai-Wei Chang, Song-Chun Zhu, Oyvind Tafjord, Peter Clark, and Ashwin Kalyan.
\newblock Learn to explain: Multimodal reasoning via thought chains for science question answering.
\newblock In \emph{The 36th Conference on Neural Information Processing Systems (NeurIPS)}, 2022.

\bibitem[Lu et~al.(2024)Lu, Bansal, Xia, Liu, Li, Hajishirzi, Cheng, Chang, Galley, and Gao]{lu2024mathvista}
Pan Lu, Hritik Bansal, Tony Xia, Jiacheng Liu, Chunyuan Li, Hannaneh Hajishirzi, Hao Cheng, Kai-Wei Chang, Michel Galley, and Jianfeng Gao.
\newblock Mathvista: Evaluating mathematical reasoning of foundation models in visual contexts.
\newblock In \emph{International Conference on Learning Representations (ICLR)}, 2024.

\bibitem[Meng et~al.(2024)Meng, Xia, and Chen]{meng2024simpo}
Yu Meng, Mengzhou Xia, and Danqi Chen.
\newblock Simpo: Simple preference optimization with a reference-free reward.
\newblock \emph{arXiv preprint arXiv:2405.14734}, 2024.

\bibitem[Oquab et~al.(2023)Oquab, Darcet, Moutakanni, Vo, Szafraniec, Khalidov, Fernandez, Haziza, Massa, El-Nouby, et~al.]{oquab2023dinov2}
Maxime Oquab, Timoth{\'e}e Darcet, Th{\'e}o Moutakanni, Huy Vo, Marc Szafraniec, Vasil Khalidov, Pierre Fernandez, Daniel Haziza, Francisco Massa, Alaaeldin El-Nouby, et~al.
\newblock Dinov2: Learning robust visual features without supervision.
\newblock \emph{arXiv preprint arXiv:2304.07193}, 2023.

\bibitem[Ouyang et~al.(2022)Ouyang, Wu, Jiang, Almeida, Wainwright, Mishkin, Zhang, Agarwal, Slama, Ray, et~al.]{instructgpt}
Long Ouyang, Jeffrey Wu, Xu Jiang, Diogo Almeida, Carroll Wainwright, Pamela Mishkin, Chong Zhang, Sandhini Agarwal, Katarina Slama, Alex Ray, et~al.
\newblock Training language models to follow instructions with human feedback.
\newblock \emph{Advances in neural information processing systems}, 35:\penalty0 27730--27744, 2022.

\bibitem[Radford et~al.(2021)Radford, Kim, Hallacy, Ramesh, Goh, Agarwal, Sastry, Askell, Mishkin, Clark, et~al.]{clip}
Alec Radford, Jong~Wook Kim, Chris Hallacy, Aditya Ramesh, Gabriel Goh, Sandhini Agarwal, Girish Sastry, Amanda Askell, Pamela Mishkin, Jack Clark, et~al.
\newblock Learning transferable visual models from natural language supervision.
\newblock In \emph{International conference on machine learning}, pages 8748--8763. PMLR, 2021.

\bibitem[Rafailov et~al.(2024)Rafailov, Sharma, Mitchell, Manning, Ermon, and Finn]{rafailov2024direct}
Rafael Rafailov, Archit Sharma, Eric Mitchell, Christopher~D Manning, Stefano Ermon, and Chelsea Finn.
\newblock Direct preference optimization: Your language model is secretly a reward model.
\newblock \emph{Advances in Neural Information Processing Systems}, 36, 2024.

\bibitem[Schulman et~al.(2017)Schulman, Wolski, Dhariwal, Radford, and Klimov]{ppo}
John Schulman, Filip Wolski, Prafulla Dhariwal, Alec Radford, and Oleg Klimov.
\newblock Proximal policy optimization algorithms.
\newblock \emph{arXiv preprint arXiv:1707.06347}, 2017.

\bibitem[Sun et~al.(2023)Sun, Shen, Cao, Liu, Li, Shen, Gan, Gui, Wang, Yang, et~al.]{llava_rlhf}
Zhiqing Sun, Sheng Shen, Shengcao Cao, Haotian Liu, Chunyuan Li, Yikang Shen, Chuang Gan, Liang-Yan Gui, Yu-Xiong Wang, Yiming Yang, et~al.
\newblock Aligning large multimodal models with factually augmented rlhf.
\newblock \emph{arXiv preprint arXiv:2309.14525}, 2023.

\bibitem[Team et~al.(2023)Team, Anil, Borgeaud, Alayrac, Yu, Soricut, Schalkwyk, Dai, Hauth, Millican, et~al.]{team2023gemini}
Gemini Team, Rohan Anil, Sebastian Borgeaud, Jean-Baptiste Alayrac, Jiahui Yu, Radu Soricut, Johan Schalkwyk, Andrew~M Dai, Anja Hauth, Katie Millican, et~al.
\newblock Gemini: a family of highly capable multimodal models.
\newblock \emph{arXiv preprint arXiv:2312.11805}, 2023.

\bibitem[Tong et~al.(2024)Tong, Liu, Zhai, Ma, LeCun, and Xie]{mmvp}
Shengbang Tong, Zhuang Liu, Yuexiang Zhai, Yi Ma, Yann LeCun, and Saining Xie.
\newblock Eyes wide shut? exploring the visual shortcomings of multimodal llms.
\newblock In \emph{Proceedings of the IEEE/CVF Conference on Computer Vision and Pattern Recognition}, pages 9568--9578, 2024.

\bibitem[Wang et~al.(2023)Wang, Wang, Xu, Zhang, Gu, Jia, Yan, Zhang, and Sang]{amber}
Junyang Wang, Yuhang Wang, Guohai Xu, Jing Zhang, Yukai Gu, Haitao Jia, Ming Yan, Ji Zhang, and Jitao Sang.
\newblock An llm-free multi-dimensional benchmark for mllms hallucination evaluation.
\newblock \emph{arXiv preprint arXiv:2311.07397}, 2023.

\bibitem[Wang et~al.(2024{\natexlab{a}})Wang, Chen, Wang, Zhou, Zhou, Yao, Zhou, Goldstein, Bhatia, Huang, and Xiao]{sima}
Xiyao Wang, Jiuhai Chen, Zhaoyang Wang, Yuhang Zhou, Yiyang Zhou, Huaxiu Yao, Tianyi Zhou, Tom Goldstein, Parminder Bhatia, Furong Huang, and Cao Xiao.
\newblock Enhancing visual-language modality alignment in large vision language models via self-improvement.
\newblock \emph{arXiv preprint arXiv:2405.15973}, 2024{\natexlab{a}}.

\bibitem[Wang et~al.(2024{\natexlab{b}})Wang, Zhang, Luo, Sun, Cui, Wang, Zhang, Wang, Li, Yu, et~al.]{wang2024emu3}
Xinlong Wang, Xiaosong Zhang, Zhengxiong Luo, Quan Sun, Yufeng Cui, Jinsheng Wang, Fan Zhang, Yueze Wang, Zhen Li, Qiying Yu, et~al.
\newblock Emu3: Next-token prediction is all you need.
\newblock \emph{arXiv preprint arXiv:2409.18869}, 2024{\natexlab{b}}.

\bibitem[Wei et~al.(2022)Wei, Wang, Schuurmans, Bosma, Xia, Chi, Le, Zhou, et~al.]{wei2022chain}
Jason Wei, Xuezhi Wang, Dale Schuurmans, Maarten Bosma, Fei Xia, Ed Chi, Quoc~V Le, Denny Zhou, et~al.
\newblock Chain-of-thought prompting elicits reasoning in large language models.
\newblock \emph{Advances in neural information processing systems}, 35:\penalty0 24824--24837, 2022.

\bibitem[Xiong et~al.(2024)Xiong, Wang, Guo, Ye, Fan, Gu, Huang, and Li]{llava_critic}
Tianyi Xiong, Xiyao Wang, Dong Guo, Qinghao Ye, Haoqi Fan, Quanquan Gu, Heng Huang, and Chunyuan Li.
\newblock Llava-critic: Learning to evaluate multimodal models.
\newblock \emph{arXiv preprint arXiv:2410.02712}, 2024.

\bibitem[Yan et~al.(2024)Yan, Bai, Chen, Zhou, Huang, and Li]{yan2024vigor}
Siming Yan, Min Bai, Weifeng Chen, Xiong Zhou, Qixing Huang, and Li~Erran Li.
\newblock Vigor: Improving visual grounding of large vision language models with fine-grained reward modeling.
\newblock \emph{arXiv preprint arXiv:2402.06118}, 2024.

\bibitem[Yao et~al.(2024)Yao, Yu, Zhang, Wang, Cui, Zhu, Cai, Li, Zhao, He, et~al.]{yao2024minicpm}
Yuan Yao, Tianyu Yu, Ao Zhang, Chongyi Wang, Junbo Cui, Hongji Zhu, Tianchi Cai, Haoyu Li, Weilin Zhao, Zhihui He, et~al.
\newblock Minicpm-v: A gpt-4v level mllm on your phone.
\newblock \emph{arXiv preprint arXiv:2408.01800}, 2024.

\bibitem[Yu et~al.(2024{\natexlab{a}})Yu, Yao, Zhang, He, Han, Cui, Hu, Liu, Zheng, Sun, et~al.]{yu2024rlhfv}
Tianyu Yu, Yuan Yao, Haoye Zhang, Taiwen He, Yifeng Han, Ganqu Cui, Jinyi Hu, Zhiyuan Liu, Hai-Tao Zheng, Maosong Sun, et~al.
\newblock Rlhf-v: Towards trustworthy mllms via behavior alignment from fine-grained correctional human feedback.
\newblock In \emph{Proceedings of the IEEE/CVF Conference on Computer Vision and Pattern Recognition}, pages 13807--13816, 2024{\natexlab{a}}.

\bibitem[Yu et~al.(2024{\natexlab{b}})Yu, Zhang, Yao, Dang, Chen, Lu, Cui, He, Liu, Chua, et~al.]{yu2024rlaif}
Tianyu Yu, Haoye Zhang, Yuan Yao, Yunkai Dang, Da Chen, Xiaoman Lu, Ganqu Cui, Taiwen He, Zhiyuan Liu, Tat-Seng Chua, et~al.
\newblock Rlaif-v: Aligning mllms through open-source ai feedback for super gpt-4v trustworthiness.
\newblock \emph{arXiv preprint arXiv:2405.17220}, 2024{\natexlab{b}}.

\bibitem[Yuan et~al.(2024)Yuan, Pang, Cho, Sukhbaatar, Xu, and Weston]{yuan2024self}
Weizhe Yuan, Richard~Yuanzhe Pang, Kyunghyun Cho, Sainbayar Sukhbaatar, Jing Xu, and Jason Weston.
\newblock Self-rewarding language models.
\newblock \emph{arXiv preprint arXiv:2401.10020}, 2024.

\bibitem[Zhang et~al.(2023{\natexlab{a}})Zhang, Wu, Zhao, Lin, Zhang, Wang, and Xie]{zhang2023pmcvqa}
Xiaoman Zhang, Chaoyi Wu, Ziheng Zhao, Weixiong Lin, Ya Zhang, Yanfeng Wang, and Weidi Xie.
\newblock Pmc-vqa: Visual instruction tuning for medical visual question answering.
\newblock \emph{arXiv preprint arXiv:2305.10415}, 2023{\natexlab{a}}.

\bibitem[Zhang et~al.(2023{\natexlab{b}})Zhang, Zhang, Gu, Zhou, Lipka, Yang, and Sun]{zhang2023llavar}
Yanzhe Zhang, Ruiyi Zhang, Jiuxiang Gu, Yufan Zhou, Nedim Lipka, Diyi Yang, and Tong Sun.
\newblock Llavar: Enhanced visual instruction tuning for text-rich image understanding.
\newblock \emph{arXiv preprint arXiv:2306.17107}, 2023{\natexlab{b}}.

\bibitem[Zhao et~al.(2023)Zhao, Wu, He, and Huang]{zhao2023svit}
Bo Zhao, Boya Wu, Muyang He, and Tiejun Huang.
\newblock Svit: Scaling up visual instruction tuning.
\newblock \emph{arXiv preprint arXiv:2307.04087}, 2023.

\bibitem[Zhou et~al.(2024{\natexlab{a}})Zhou, Cui, Rafailov, Finn, and Yao]{povid}
Yiyang Zhou, Chenhang Cui, Rafael Rafailov, Chelsea Finn, and Huaxiu Yao.
\newblock Aligning modalities in vision large language models via preference fine-tuning.
\newblock \emph{arXiv preprint arXiv:2402.11411}, 2024{\natexlab{a}}.

\bibitem[Zhou et~al.(2024{\natexlab{b}})Zhou, Fan, Cheng, Yang, Chen, Cui, Wang, Li, Zhang, and Yao]{csr}
Yiyang Zhou, Zhiyuan Fan, Dongjie Cheng, Sihan Yang, Zhaorun Chen, Chenhang Cui, Xiyao Wang, Yun Li, Linjun Zhang, and Huaxiu Yao.
\newblock Calibrated self-rewarding vision language models.
\newblock \emph{arXiv preprint arXiv:2405.14622}, 2024{\natexlab{b}}.

\bibitem[Zhu et~al.(2023)Zhu, Chen, Shen, Li, and Elhoseiny]{zhu2023minigpt}
Deyao Zhu, Jun Chen, Xiaoqian Shen, Xiang Li, and Mohamed Elhoseiny.
\newblock Minigpt-4: Enhancing vision-language understanding with advanced large language models.
\newblock \emph{arXiv preprint arXiv:2304.10592}, 2023.

\end{thebibliography}
}

\clearpage
\setcounter{page}{1}
\maketitlesupplementary

\begin{table}[h]
    \centering
    \resizebox{\linewidth}{!}{
    \begin{tabular}{l|lll}
    \toprule
    Parameter   & Critic  Training& DPO Training & Enhanced SFT \\
    \midrule
    Regularization $\alpha$ & - & 1e-3 & - \\
    Lora r   & 128 &128&128\\
    Lora $\alpha$ & 256 &256&256\\
    Mm\_projector lr & 2e-5 & 2e-5&2e-5\\
    Learning rate & 2e-5 &1e-7&2e-5\\
    Total batch\_size & 32 &1&32\\
    epoch &1&1&1\\
    optimizer& AdamW& AdamW& AdamW\\
    Deepspeed Stage&2&2&2 \\

    \bottomrule
    \end{tabular}
    }
    \caption{The hyperparameter settings of training. }
    \label{tab:train_detail}
\end{table}

\section{The Comparison of Critics}
\label{examples}

Here, we present some examples of the comparison of critics from different models in Table \ref{tab:critic_example_1}, \ref{tab:critic_example_2}, \ref{tab:critic_example_3}, and \ref{tab:critic_example_4}. GPT-4o and our Critic model produce similar scores for responses, but they fail to identify the flaws in bad responses from the baseline LLAVA model.

\begin{table*}[!htbp]
\begin{minipage}{1.0\textwidth}
\centering
\small
\scalebox{0.9}{
\begin{tabular}{l p{16cm}}
\toprule
 \multicolumn{2}{l}{\bf Example 1: The comparison of critic from multiple models}  \\
\midrule
&  \includegraphics[height=2.7cm]{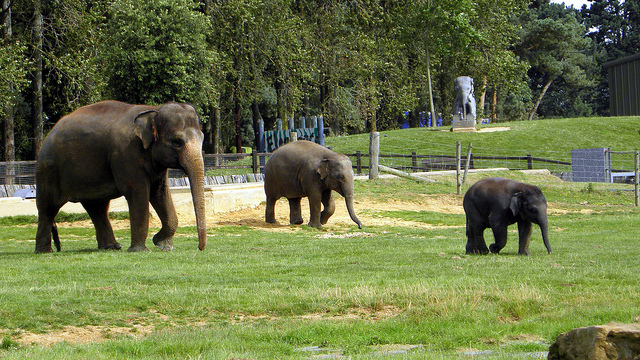} \\

& \textbf{Question:} Summarize the visual content
presented. \\
& \textbf{Response 1}: The image shows a group of elephants in a grassy enclosure. There are three elephants visible: one large adult elephant, one medium-sized juvenile elephant, and one smaller baby elephant. The adult elephant is on the left side of the image, the juvenile is in the middle, and the baby is on the right. They are standing on a grassy area with some trees in the background. The enclosure appears to be a part of a zoo or wildlife park, as indicated by the fencing and the presence of a statue in the background. \\
& \textbf{System Message}: Critic Prompt \\
\midrule
\textbf{GPT-4o}  & \textcolor{green}{5 4 5 4 4 22} \\
&Assessment based on the response:
Relevance (5): The response accurately describes the image content with direct relevance to the user's inquiry.
Substantial Coverage (4): The response covers most aspects of the visual content, including the size and position of the elephants and the setting, but doesn't delve into further details or context.
Basic Elements (5): The response provides the basic description elements, such as the number, size, and arrangement of the elephants and the presence of a fenced enclosure.
Clarity and Organization (4): The response is clear, organized, and easy to follow, but minor improvements could enhance conciseness.
High Quality (4): The response is well-tailored to the question, though it could include more expert-level insights or contextual analysis.
 \\
\midrule
\textbf{Our Critic Model} & \textcolor{green}{5 5 5 5 5 25} \\
 &  Here's the assessment for each criterion:
1. Relevance: The response directly relates to the user's request by accurately summarizing the visual content of the image, focusing on the elephants and their environment. Score: 5
2. Substantial Coverage: It comprehensively addresses the scene depicted in the image, noting the number and types of elephants, as well as contextual elements like the enclosure and background. Score: 5
3. Basic Elements: The response captures all essential details such as the classification of the elephants by age and size, and describes the setting effectively. Score: 5
4. Clarity and Organization: The summary is well-organized and presented in a logical sequence, making it easy to understand. The information is clear and directly relevant to the image content. Score: 5
5. High Quality: The response is free of errors, concise, and focuses precisely on the task of summarizing the image without deviating into irrelevant details. Score: 5
Total score: 25
\\
 \midrule
LLaVA-v1.6 7B~\cite{liu2023improvedllava}&  5 5 5 5 5 25 \\
 & Relevance: The response effectively describes the image content and remains relevant to the user's inquiry. Score: 5
Substantial Coverage: The response captures most aspects of the visual content, including the elephants, their positions, and the zoo setting, offering a thorough overview. Score: 5
Basic Elements: The response includes all key descriptive elements, such as the number and size of the elephants, their arrangement, and the enclosure details. Score: 5
Clarity and Organization: The response is well-organized, clear, and easy to follow, providing a structured and concise summary of the image. Score: 5
High Quality: The response is comprehensive and tailored to the user's query, demonstrating expert-level insight and precision in its description. Score: 5 \\
\bottomrule
\end{tabular}
}
\vspace{0mm}
\captionof{table}{The analysis of three different models—GPT-4o, a proprietary Critic Model, and LLaVA-v1.6—based on one good response to summarizing an image depicting elephants.}
\label{tab:critic_example_1}  
\end{minipage}
\end{table*}

\begin{table*}[!htbp]
\begin{minipage}{1.0\textwidth}
\centering
\small
\scalebox{0.9}{
\begin{tabular}{l p{16cm}}
\toprule
 \multicolumn{2}{l}{\bf Example 2: The comparison of critic from multiple models}  \\
\midrule
&  \includegraphics[height=2.7cm]{figs/COCO_train2014_000000571441.jpg} \\

& \textbf{Question:} Summarize the visual content
presented. \\
& \textbf{Response 2}: The image shows a group of elephants in an enclosure. There are three elephants visible: one adult elephant and two younger ones. The adult elephant is standing on the left side of the image, while \textcolor{red}{the two younger elephants are on the right}. The enclosure has a fence and there are some structures that could be part of the enclosure's infrastructure. The setting appears to be a zoo or wildlife park. \textcolor{red}{The sky is overcast, suggesting it might be a cloudy day}. \\
& \textbf{System Message}: Critic Prompt \\
\midrule
\textbf{GPT-4o}  & \textcolor{red}{3 3 3 4 2 15} \\
 & Relevance (3): The response is relevant to the user's inquiry, describing the elephants and the enclosure. \textcolor{red}{However, inaccuracies, such as the placement of the younger elephants and the description of the sky, reduce its relevance.}
Substantial Coverage (3): The response addresses the main elements of the image, such as the presence of elephants and the zoo setting, \textcolor{red}{but the inclusion of incorrect details undermines the comprehensiveness of the coverage}.
Basic Elements (3): The response captures some basic elements of the image but fails to accurately describe the positions of the elephants and \textcolor{red}{introduces an incorrect description of the sky}.
Clarity and Organization (4): The response is well-structured and easy to follow, though the inaccuracies affect its overall clarity and utility.
High Quality (2): The response lacks precision and expert-level insight due to the inclusion of incorrect details and missed opportunities to provide deeper context or analysis.

score: 15
 \\
\midrule
\textbf{Our Critic Model}  & \textcolor{red}{5 2 2 5 4 18} \\
 & Here's the assessment for each criterion based on the provided response:
1. Relevance: The response is relevant as it accurately describes the elephants, their placement within the image, and the setting. Score: 5
2. Substantial Coverage: The response covers the major visual elements of the image, including the composition and number of elephants, and hints at the environment. \textcolor{red}{However, it lacks the detail of the juvenile elephants' specific sizes and the presence of the statue in the background}. Score: 2
3. Basic Elements: It identifies the basic elements like the number of elephants and their approximate ages, as well as the setting of a zoo or park, \textcolor{red}{but misses some specifics}. Score: 2
4. Clarity and Organization: The response is clear and straightforward, describing the elements of the image in a logical sequence. Score: 5
5. High Quality: The summary is concise and relevant but lacks some details that would provide a fuller understanding of the visual scene. Score: 4
Total score: 18 \\
 \midrule
LLaVA-1.6 7B~\cite{liu2023improvedllava}&  5 5 5 5 5 25 \\
 & Relevance: The response effectively describes the image content and remains relevant to the user's inquiry. Score: 5
Substantial Coverage: The response captures most aspects of the visual content, including the elephants, their positions, and the zoo setting, offering a thorough overview. Score: 5
Basic Elements: The response includes all key descriptive elements, such as the number and size of the elephants, their arrangement, and the enclosure details. Score: 5
Clarity and Organization: The response is well-organized, clear, and easy to follow, providing a structured and concise summary of the image. Score: 5
High Quality: The response is comprehensive and tailored to the user's query, demonstrating expert-level insight and precision in its description. Score: 5 \\
\bottomrule
\end{tabular}
}
\vspace{0mm}
\captionof{table}{Evaluation of three models—GPT-4o, a proprietary Critic Model, and LLaVA-v1.6-7B based on their ability to assess a flawed image summary involving elephants. GPT-4o and our Critic Model successfully identified errors in the caption, whereas LLaVA-v1.6-7B failed to detect discrepancies between the image and the generated caption.}
\label{tab:critic_example_2}  
\end{minipage}
\end{table*}

\begin{table*}[!htbp]
\begin{minipage}{1.0\textwidth}
\centering
\small
\scalebox{0.9}{
\begin{tabular}{l p{16cm}}
\toprule
 \multicolumn{2}{l}{\bf Example 3: The comparison of critic from multiple models}  \\
\midrule
&  \includegraphics[height=2.7cm]{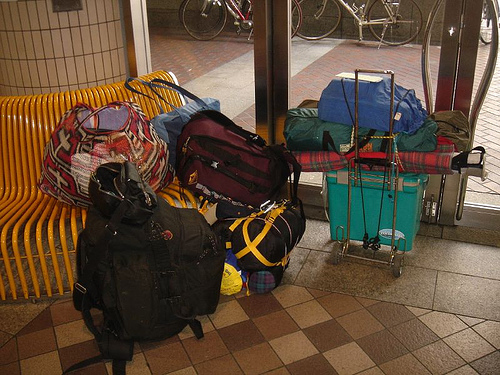} \\

& \textbf{Question:} Summarize the visual content
presented. \\
& \textbf{Response 1}: The image shows a collection of luggage and personal items arranged on a bench and a luggage cart. The luggage includes various types of bags and suitcases, some of which are strapped together. The items are likely belongings of someone who is either arriving or departing from a location, possibly a train station or a similar public transportation hub. The setting appears to be an indoor waiting area, as indicated by the presence of a bench, a luggage cart, and a door in the background. The floor is tiled, and there are two bicycles parked outside the door. \\
& \textbf{System Message}: Critic Prompt \\
\midrule
\textbf{GPT-4o}  & \textcolor{green}{5 5 5 5 5 25} \\
 &Relevance (5): The response is highly relevant as it accurately describes the primary content of the image, focusing on the luggage, setting, and other visible elements.
Substantial Coverage (5): The response covers a substantial portion of the image details, including the description of luggage types, the setting, and additional elements like the bicycles outside and the tiled floor.
Basic Elements (5): The response provides the basic elements of the image, such as the luggage arrangement, the bench, the luggage cart, and the setting, making it complete and useful.
Clarity and Organization (5): The response is clearly written, well-organized, and provides a concise yet comprehensive summary of the image content.
High Quality (5): The response is well-tailored to the image, offering detailed and insightful observations without unnecessary information.

score: 25 
 \\
\midrule
\textbf{Our Critic Model}  & \textcolor{green}{5 5 5 5 5 25} \\
 & Here's the assessment for each criterion:
Relevance: The response directly addresses the user's instruction to summarize the visual content of the image, describing the collection of luggage and its probable setting. Score: 5.
Substantial Coverage: The response provides a substantial summary of the visual information, noting the types of luggage, their arrangement, and suggesting a likely location. Score: 5.
Basic Elements: The response covers the basic elements present in the image such as types of luggage, and environmental details like the tiled floor and bicycles, which are key in constructing a clear visual summary. Score: 5.
Clarity and Organization: The summary is clear, well-organized, and presents the visual information in a logical manner. It clearly distinguishes different elements in the image and their potential relevance. Score: 5.
High Quality: The response is tailored to provide a detailed and comprehensive summary of the image without extraneous information, reflecting a high-quality analysis suitable for the user's request. Score: 5.
Total score: 25 \\
 \midrule
LLaVA-1.6 7B~\cite{liu2023improvedllava}&  5 5 5 5 5 25 \\
 &  Relevance: The response is highly relevant, accurately describing the content of the image and focusing on the luggage, setting, and visible elements like the bicycles outside. Score: 5.
Substantial Coverage: The response covers all significant aspects of the image, including the description of the luggage, the indoor setting, and contextual details such as the bench and the bicycles. Score: 5.
Basic Elements: The response includes all essential elements, such as the luggage arrangement, the bench, the luggage cart, and the tiled floor, providing a complete and useful summary. Score: 5.
Clarity and Organization: The response is clearly written, well-structured, and effectively communicates the details of the image in a concise and coherent manner. Score: 5.
High Quality: The response is well-tailored to the image, demonstrating expert-level insight and attention to detail without including irrelevant information. Score: 5.
score: 25\\
\bottomrule
\end{tabular}
}
\vspace{0mm}
\captionof{table}{The analysis of three different models—GPT-4o, a proprietary Critic Model, and LLaVA-v1.6—based on one good response to summarizing an image showing luggage.}
\label{tab:critic_example_3}  
\end{minipage}
\end{table*}

\begin{table*}[!htbp]
\begin{minipage}{1.0\textwidth}
\centering
\small
\scalebox{0.9}{
\begin{tabular}{l p{16cm}}
\toprule
 \multicolumn{2}{l}{\bf Example 4: The comparison of critic from multiple models}  \\
\midrule
&  \includegraphics[height=2.7cm]{figs/COCO_train2014_000000140883.jpg} \\

& \textbf{Question:} Summarize the visual content
presented. \\
& \textbf{Response 2}: The image shows a collection of luggage and bags piled up against a wall, with a bench and \textcolor{red}{a bicycle} in the background. The luggage appears to be a mix of backpacks, duffel bags, and suitcases, with various colors and designs. The setting seems to be an indoor public space, possibly a waiting area or a transit hub. \\
& \textbf{System Message}: Critic Prompt \\
\midrule
\textbf{GPT-4o}  & \textcolor{red}{4 4 4 5 3 20} \\
 & Relevance (4): The response is generally relevant to the image but inaccurately mentions "a bicycle" in the background, which reduces its alignment with the visual content.
Substantial Coverage (4): The response captures the overall context, such as the luggage, the indoor setting, and the mix of bags. However, it omits details like the luggage cart and the bicycles outside the door, which are present in the image.
Basic Elements (4): The response provides most of the basic elements but includes an incorrect detail (the bicycle being indoors) and misses some key elements like the tiled floor and luggage arrangement.
Clarity and Organization (5): The response is clear and well-organized, making it easy to understand despite the inaccuracies.
High Quality (3): The response is moderately insightful but \textcolor{red}{lacks precision due to the incorrect mention of a bicycle indoors and the omission of significant details}.
score: 20
 \\
\midrule
\textbf{Our Critic Model}  & \textcolor{red}{2 5 2 2 3 14} \\
 & Here's the assessment for each criterion based on the provided response: Relevance: The response is relevant to request for a summary of the visual content in the image, \textcolor{red}{wrongly identifying key elements like two bicycles}. Score: 2.
Substantial Coverage: The response covers a significant part of the visual information by describing the types of luggage and the setting. But it could provide more detail about the arrangement of items. Score: 5.
Basic Elements: The response identifies basic elements such as luggage types and the background setting \textcolor{red}{but lacks specific details about the arrangement and other items, which are also visually significant}. Score: 2.
Clarity and Organization: \textcolor{red}{The omission of the door with a glass panel affects the overall clarity and accuracy of the description}, particularly in defining the specific features of the setting. Score: 2.
High Quality: The response is of good quality, providing a useful summary of the image. But for an impeccable response, more precise descriptions and inclusion of all relevant visual elements are needed. Score: 3.
Total score: 14\\
 \midrule
LLaVA-1.6 7B~\cite{liu2023improvedllava}&  5 5 5 5 5 25 \\
 & Relevance: The response is highly relevant, focusing on the luggage and the indoor setting, aligning with the main elements of the image. Score: 5.
Substantial Coverag: The response effectively captures the primary aspects of the image, such as the variety of luggage and the setting as a public indoor space, providing sufficient context. Score: 5.
Basic Elements: The response includes the key descriptive elements, such as the types of bags, their arrangement, and the setting, creating a comprehensive overview. Score: 5.
Clarity and Organization: The response is clearly written, well-structured, and easy to understand, maintaining a logical flow of information. Score: 5.
High Quality: The response is expertly tailored, providing an insightful description that aligns well with the visual content, despite not explicitly mentioning all minor details. Score: 5.

score: 25 \\
\bottomrule
\end{tabular}
}
\vspace{0mm}
\captionof{table}{Comparison of three models—GPT-4o, a proprietary Critic Model, and LLaVA-v1.6-7B in evaluating a flawed response summarizing an image of luggage. Both GPT-4o and our Critic Model successfully identified errors in the caption, while LLaVA-v1.6-7B failed to detect the hallucination between the image and the caption.}
\label{tab:critic_example_4}  
\end{minipage}
\end{table*}

\section{Critic Prompts}
\label{cri_prompt}

In the experiment, we utilize prompts in three different styles. As shown in Table \ref{cri_prompt}, most of the experiment is conducted with prompts in rating style, apart from the ablation study presented in Section \ref{ablation}.

\begin{table*}[!htbp]
\begin{minipage}{1.0\textwidth}
\centering
\small
\scalebox{0.9}{
\begin{tabular}{l p{16cm}}
\toprule
 \multicolumn{2}{l}{\bf The Critic Prompts in Different Styles}  \\
\midrule
\textbf{Rating Style}  & Review the user’s question and the corresponding response using these criteria. Points are accumulated
based on the satisfaction of each
criterion:

- \textcolor{red}{Relevance}: Is the response relevant and provides some information related to the user’s inquiry and
visual information, even if it is incomplete or contains some irrelevant content?

- \textcolor{red}{Substantial Coverage}: if the response addresses a substantial portion of the user’s question and visual
information but does not completely resolve the query or provide a direct answer.

- \textcolor{red}{Basic Elements}: if the response answers the basic elements of the user’s question and visual information
in a useful way, regardless of whether it seems to have been written by an AI Assistant or if it has
elements typically found in blogs or search results.

- \textcolor{red}{Clarity and Organization}: if the response is clearly written from an AI Assistant’s perspective,
addressing the user’s question directly and summarize the visual information comprehensively, and is
well-organized and helpful, even if there is slight room for improvement in clarity, conciseness or focus.

- \textcolor{red}{High Quality}: for a response that is impeccably tailored to the user’s question by an AI Assistant,
without extraneous information, reflecting expert knowledge, and demonstrating a high-quality, engaging,
and insightful answer.

After examining the user’s instruction and the response:
User’s instruction: \textcolor{red}{$\langle$ Instruction $\rangle$}

The assistant’s Response: \textcolor{red}{$\langle$Response$\rangle$}

Provide a concise assessment with a score from 1 to 5 for each criterion, and the scores of these criteria should be additive for a total score. Conclude with the score using the format: “score:  $\langle$total points $\rangle$” \\
\midrule
\textbf{Additive Style}~\cite{yuan2024self} & Review the user’s question and the corresponding response using these criteria. Points are accumulated
based on the satisfaction of each
criterion:

- Add 1 point if the response is relevant and provides some information related to the user’s inquiry and
visual information, even if it is incomplete or contains some irrelevant content.

- Award another point if the response addresses a substantial portion of the user’s question and visual
information but does not completely resolve the query or provide a direct answer.

- Give a third point if the response answers the basic elements of the user’s question and visual information in a useful way, regardless of whether it seems to have been written by an AI Assistant or if it has elements typically found in blogs or search results.

- Award a fourth point if the response is clearly written from an AI Assistant’s perspective, addressing the user’s question directly and summarize the visual information comprehensively, and is well-organized and helpful, even if there is slight room for improvement in clarity, conciseness or focus.

- Add a fifth point for a response that is impeccably tailored to the user’s question by an AI Assistant, without extraneous information, reflecting expert knowledge, and demonstrating a high-quality, engaging, and insightful answer.

After examining the user’s instruction and the response:
User’s instruction: \textcolor{red}{$\langle$ Instruction $\rangle$}

The assistant’s Response: \textcolor{red}{$\langle$Response$\rangle$}

Provide a concise assessment with a score from 1 to 5 for each criterion, and the scores of these criteria should be additive for a total score. Conclude with the score using the format: “score: $\langle$total points $\rangle$”\\
 \midrule
\textbf{Subtractive Style}&  Review the user’s question and the corresponding response using the subtractive 5-point scoring systemdescribed below. The initial point is 5. Points are deducted based on the satisfaction of each criterion:

- Deduct 1 point if the response contains some information related to the user’s question or visual information, but it’s incomplete or has some irrelevant content.

- Deduct another point if the response addresses only a substantial portion of the user’s question and visual information, but doesn’t fully resolve the query or provide a direct answer.

- Subtract a third point if the response answers the basic elements of the user’s question and visual information in a useful way, even if it has elements typically found in AI-generated or blog-like responses.

- Remove a fourth point if the response is clearly written from an AI Assistant’s perspective, directly addressing the user’s question and summarizing the visual information comprehensively, while being well-organized and helpful but with slight room for improvement in clarity, conciseness, or focus.

- Deduct a fifth point if the response is impeccably tailored to the user’s question, without extraneous
information, reflecting expert knowledge, and demonstrating a high-quality, engaging, and insightful
answer. 

After examining the user’s instruction and the response:
User’s instruction: \textcolor{red}{$\langle$ Instruction $\rangle$}

The assistant’s Response: \textcolor{red}{$\langle$Response$\rangle$}

Provide a concise assessment with a score from 1 to 5 for each criterion, and the scores of these criteria should be additive for a total score. Conclude with the score using the format: “score: $\langle$ total points$\rangle$”\\
\bottomrule
\end{tabular}
}
\vspace{0mm}
\captionof{table}{The prompts in three different styles.}
\label{tab:critic_prompts}  
\end{minipage}
\end{table*}


    


\section{Training Details}
\label{train_details}

The training comprises the Critic model training, DPO training, and enhanced SFT. The training details are shown in Table \ref{tab:train_detail}.






\section{Baseline Details}
\label{baseline_details}

\textbf{LLaVA-RLHF}~\cite{llava_rlhf} proposes a novel alignment algorithm called Factually Augmented RLHF that enhances the reward model by incorporating additional factual data, such as image captions and ground-truth multiple-choice options. Using annotated preference data, one round of preference learning is conducted on LLaVA1.5.

\textbf{RLHF-V}~\cite{yu2024rlhfv} collects fine-grained, paragraph-level corrections from humans on hallucinations and performs dense Direct Preference Optimization (DPO) based on this human feedback, targeting specific areas where hallucinations occur.

\textbf{Silkie}~\cite{2023vlfeedback} constructs the VLFeedback dataset using annotations from Vision-Language Large Models (VLLMs). Responses are generated by 12 LVLMs conditioned on multimodal instructions from various datasets, and the outputs are evaluated by GPT-4V for helpfulness, visual faithfulness, and ethical considerations. This dataset is used to perform DPO on LLaVA-1.5.

\textbf{POVID}~\cite{povid} introduces a novel training paradigm that aligns preferences in VLLMs by utilizing external preference data from GPT-4 and exploiting the model’s inherent hallucination patterns triggered by noisy images.

\textbf{SIMA}~\cite{sima} leverages prompts from existing vision instruction-tuning datasets to self-generate responses and uses an in-context self-critic mechanism to select preferred response pairs for tuning, promoting enhanced response selection without external feedback.

\textbf{CSR}~\cite{csr} Implements a self-improvement method where the model iteratively generates candidate responses, evaluates rewards for each response, and curates preference data for fine-tuning. CSR’s reward modeling uses a step-wise strategy with visual constraints, emphasizing the importance of visual input.

\textbf{STIC}~\cite{stic} is a self-training approach focused on image comprehension. It self-constructs a preference dataset from unlabeled images, generating preferred responses with step-by-step prompts and dis-preferred responses from corrupted images or misleading prompts.

\textbf{RLAIF-V}~\cite{yu2024rlaif} uses a divide-and-conquer strategy by breaking down responses into atomic claims for more reliable assessments. In its feedback learning algorithm, RLAIF-V addresses the distribution shift issues of traditional DPO through an online learning approach, enhancing learning performance and efficiency.

\textbf{Self-Rewarding}~\cite{yuan2024self} introduces self-feedback learning in LLMs, where the model is given a critic prompt and tasked with evaluating which output is better, promoting self-directed improvement without external annotations.

\section{Benchmark Details}

\textbf{MME}~\cite{fu2023mme} serves as a comprehensive benchmark, aiming to assess perceptual and cognitive capability of MLLMs within 14 sub-tasks.

\textbf{SEEDBench}~\cite{li2023seed} is a comprehensive benchmark that evaluates the hierarchical capabilities of MLLMs, including the generation of both texts and images given interleaved image-text inputs.

\textbf{ScienceQA}~\cite{sciqa} is a benchmark that consists of more than 21k multimodal multiple-choice questions with diverse science topics and annotations of their answers, which include three subjects: natural science, language science, and social science.

\textbf{HallusionBench}~\cite{guan2023hallusionbench} is an image-context reasoning benchmark designed for the challenges of hallucination in MLLMs, which compreses 346 images paired with 1,129 questions, all meticulously crafted by human experts.

\textbf{AMBER}~\cite{amber} is a multi-dimensional hallucination benchmark comprising more than 15k samples, which can be used to evaluate both generative tasks and discriminative tasks including existence, attribute, and relation hallucination. We calculate the average accuracy as the score.

\textbf{POPE}~\cite{pope} is built on the basis of datasets with annotations about objects in the image, such as COCO. With the help of automatic segmentation tools like SEEM, it conducts negative sampling for nonexistent objects in the image under three different settings.

\textbf{MathVista}~\cite{lu2024mathvista} is designed to benchmark the MLLMs to solve diverse mathematical visual tasks. It consists of more than 6k examples from 31 datasets involving math. We utilize the \texttt{GPT-4-turbo} as the evaluator.

\section{Critic Dataset}

Here, we will present some examples of Critic Dataset for Critic Model training in Table \ref{tab:critic_data}.

\begin{table*}[!htbp]
\begin{minipage}{1.0\textwidth}
\centering
\small
\scalebox{0.9}{
\begin{tabular}{l p{16cm}}
\toprule
 \multicolumn{2}{l}{\bf Example 4: The comparison of critic from multiple models}  \\
\midrule
&  \includegraphics[height=2.7cm]{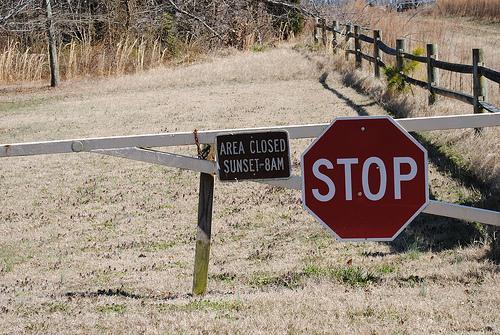} \\

& \textbf{Question:} Are there any stop signs with yellow writing on them?\\
\midrule
\textbf{Response 1}  & There is no existence of a stop sign with yellow writing in the image description. \\
 \textbf{Critic}& Helpfulness: Rationale: The response is helpful as it answers the user's question about the color of the writing on the stop sign, but it does not add any additional information that might increase the user's understanding of the image. Rating: 4
 
 Ethical Considerations: Rationale: The response is ethical, and there are no concerns regarding safety, privacy, fairness, or harmfulness. Rating: 5,
 
 Visual Faithfulness: Rationale: The response accurately reflects the image content by correctly describing the color of the stop sign. Rating: 5
 
 Total Score: 14
 \\
\midrule
\textbf{Response 2}  &No, there are no stop signs with yellow writing on them. The stop sign in the image has red writing.  \\
 \textbf{Critic}& Helpfulness: Rationale: This response not only answers the user's question but also provides valuable additional context about the stop sign's color, location, and the presence of another sign, which enhances the user's understanding of the image. Rating: 5
 
 Ethical Considerations: Rationale: This response is ethical and safe, provides accurate information, respects privacy, and does not contain any unfair or harmful content. Rating: 5
 
 Visual Faithfulness: Rationale: This response is highly faithful to the image content, correctly identifying the stop sign's colors and providing additional context about other elements in the image. Rating: 5
 
 Total Score: 15\\
 \midrule
\textbf{Response 3}&Stop signs with yellow writing indeed existed. Historically, yellow was the standard color for STOP signs for about 30 years.    \\
 \textbf{Critic} &Helpfulness: Rationale: This response is not helpful because it inaccurately affirms that there is a stop sign with yellow writing, which contradicts the visible content of the image. Additionally, it includes a lot of ungrounded and incorrect details about the image.,Rating: 1
 
 Ethical Considerations: Rationale: This response contains many inaccuracies and could potentially mislead or confuse the user, which is an ethical concern in terms of the reliability of the information provided. Rating: 1
 
 Visual Faithfulness: Rationale: This response is significantly inaccurate as it misrepresents the color of the writing on the stop sign and includes many other incorrect descriptions of the image. Rating: 1
 
 Total Score: 3  \\
\bottomrule
\end{tabular}
}
\vspace{0mm}
\captionof{table}{Examples of Critic Dataset. The 1st and 2nd responses are relevant to the question and provide a direct answer. However, the last response provides an irrelevant answer and introduces hallucination.}
\label{tab:critic_data_example}  
\end{minipage}
\end{table*}

\section{The Comparison between Models}

In the experiments, EACO achieves remarkable improvement on hallucination reduction. Here, we will show some examples between EACO and baseline LLaVA-v1.6-Mistral-7B in Table \ref{tab:amber_output_1} and \ref{tab:amber_output_2}.

\begin{table*}[!htbp]
\begin{minipage}{1.0\textwidth}
\centering
\small
\scalebox{0.9}{
\begin{tabular}{l p{14cm}}
\toprule
 \multicolumn{2}{l}{\bf Example 1: The comparison between models for Hallucination Mitigation}  \\
\midrule
&  \includegraphics[height=2.7cm]{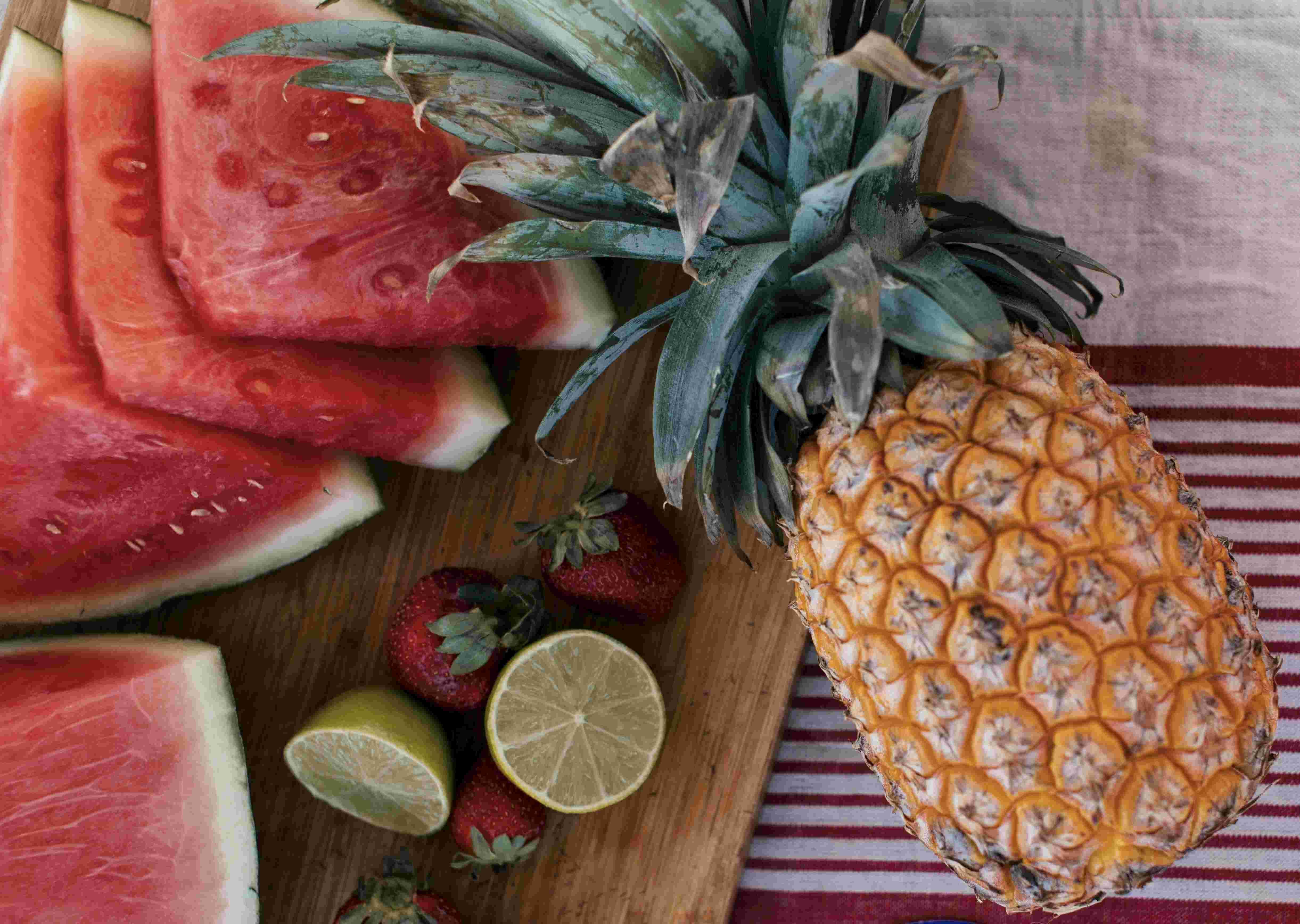} \\

& \textbf{Question:} Is there direct contact between the lemon and tablecloth?\\
\midrule
\textbf{Answer}  &  No.\\
\midrule
\textbf{EACO} (LLaVA-1.6) &No, \textcolor{green}{there is no direct contact between the lemon and the tablecloth}. The lemon is placed on a cutting board, which is on the tablecloth.  \\
 \midrule
LLaVA-v1.6 7B~\cite{liu2023improvedllava}&  Yes, \textcolor{red}{there is direct contact between the lemon and the tablecloth}. The lemon is placed on the tablecloth, which is covering the table. \\
\bottomrule
\end{tabular}
}
\vspace{0mm}
\captionof{table}{Comparison between EACO (ours) and the baseline LLaVA-v1.6-Mistral-7B. In the image, a cutting board is present between the lemon and the tablecloth. EACO successfully detects the cutting board, whereas LLaVA-v1.6-7B fails to recognize it.}
\label{tab:amber_output_1}  
\end{minipage}
\end{table*}

\begin{table*}[!htbp]
\begin{minipage}{1.0\textwidth}
\centering
\small
\scalebox{0.9}{
\begin{tabular}{l p{14cm}}
\toprule
 \multicolumn{2}{l}{\bf Example 2: The comparison between models for Hallucination Mitigation}  \\
\midrule
&  \includegraphics[height=2.7cm]{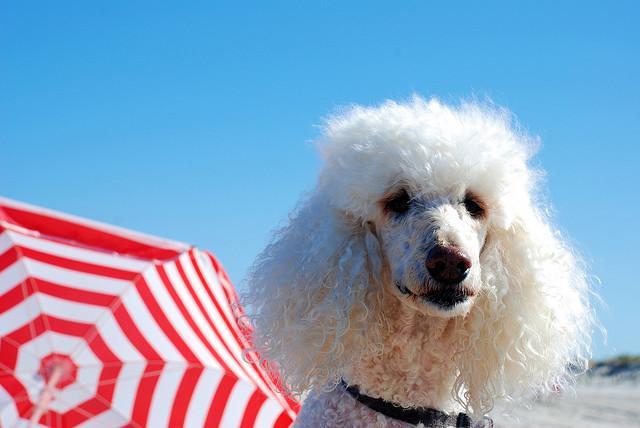} \\

& \textbf{Question:} Is there a lounge in this image? \\
\midrule
\textbf{Answer}  &No.  \\
\midrule
\textbf{EACO} (LLaVA-1.6) & \textcolor{green}{No. Based on the image provided, there is no lounge visible}. The image shows a dog with a fluffy white coat, standing next to a red and white striped umbrella. The background appears to be a beach setting with a clear blue sky. \\
 \midrule
LLaVA-v1.6 7B~\cite{liu2023improvedllava}&  \textcolor{red}{Yes}, \textcolor{red}{there is a lounge in the image}. A small white dog is sitting under a red and white striped umbrella, which is placed on a lounge chair. The dog appears to be enjoying the shade provided by the umbrella. \\
\bottomrule
\end{tabular}
}
\vspace{0mm}
\captionof{table}{Comparison between EACO (ours) and the baseline LLaVA-v1.6-Mistral-7B. In the image, no lounge is present, yet LLaVA-v1.6-7B introduces this hallucination. EACO successfully avoids this error.}
\label{tab:amber_output_2}  
\end{minipage}
\end{table*}

\end{document}